%% file: main.tex
\documentclass[a4paper, 10pt, conference]{ieeeconf}      

\IEEEoverridecommandlockouts                              
\usepackage[utf8]{inputenc}
\usepackage[T1]{fontenc}
\usepackage[table,xcdraw]{xcolor}

\usepackage{comment}
\usepackage{graphics} 
\usepackage{epsfig} 
\usepackage{hyperref}
\usepackage{multirow}
\usepackage{booktabs}
\usepackage{subfigure}

\title{\LARGE \bf
Image Classification with Classic and Deep Learning Techniques
}

\author{{
\parbox{7 in}{
    \centering
    \`Oscar Lorente Corominas \qquad
    Ian Riera Smolinska \qquad
    Aditya Sangram Singh Rana \\
    Master in Computer Vision\\
    Universitat Aut\`onoma de Barcelona\\
    08193 Bellaterra, Barcelona, Spain\\
    {\tt\small\{oscar.lorentec, 
        ianpau.riera,
        adityasangramsingh.rana\}@e-campus.uab.cat
    }}
}}

\begin{document}

\maketitle
\thispagestyle{empty}
\pagestyle{empty}

\begin{abstract}
To classify images based on their content is one of the most studied topics in the field of computer vision. Nowadays, this problem can be addressed using modern techniques such as Convolutional Neural Networks (CNN), but over the years different classical methods have been developed. In this report, we implement an image classifier using both classic computer vision and deep learning techniques. Specifically, we study the performance of a Bag of Visual Words classifier using Support Vector Machines, a Multilayer Perceptron, an existing architecture named InceptionV3 and our own CNN, TinyNet, designed from scratch. We evaluate each of the cases in terms of accuracy and loss, and we obtain results that vary between 0.6 and 0.96 depending on the model and configuration used.
\end{abstract}


\input{sections/00_intro}

\input{sections/01_problem}

\input{sections/02_data}

\input{sections/03_method}

\input{sections/conclusions}

\bibliographystyle{unsrt}
\bibliography{bibliography}
\addtolength{\textheight}{-12cm}   

\end{document}

%% file: sections/00_intro.tex
\section{INTRODUCTION}
\label{sec:intro} 
During the last years, there has been a great increase in the number of applications in which image classification is useful. Helping people organise their photo collections, analysing medical images or identifying what's around self-driving cars are just a few examples. These tasks require precisely labeled large-scale datasets, and most of them include a huge variety of image types, from dogs or cats, to landscapes, roads, and so on.

In image classification, given an input image, the goal is to predict the class which it belongs to. This is not a big deal for humans, but teaching computers to \textit{see} is a difficult problem that has become a broad area of research interest, and both classic computer vision and Deep Learning (DL) techniques have been developed. Classic techniques use local descriptors to try to find similarities between images, but today, advances in technology allow the use of DL techniques to automatically extract representative features and patterns from each image. However, to understand these concepts it is first necessary to review traditional techniques.

In this report, we present different image classification systems trained on a specific dataset detailed in Section~\ref{sec:data}. We first explore the Bag of Visual Words (BoVW) approach, which consists on extracting local features from the images and clustering them to create visual words. For each image, its histogram of visual words is used as a global feature to classify it. On the other hand, we use a Multilayer Perceptron (MLP) as a first step towards DL. To improve the performance of the system, we use CNNs, which are more suitable for image classification. Specifically, we first refine an existing architecture, InceptionV3~\cite{inception}, and finally design a CNN from scratch and analyze the impact of each layer, parameters, activation function, and more.

%% file: sections/01_problem.tex
\section{PROBLEM DEFINITION}
\label{sec:problem} 

\begin{figure}[t!]
    \centering
    \includegraphics[width=0.9\linewidth]{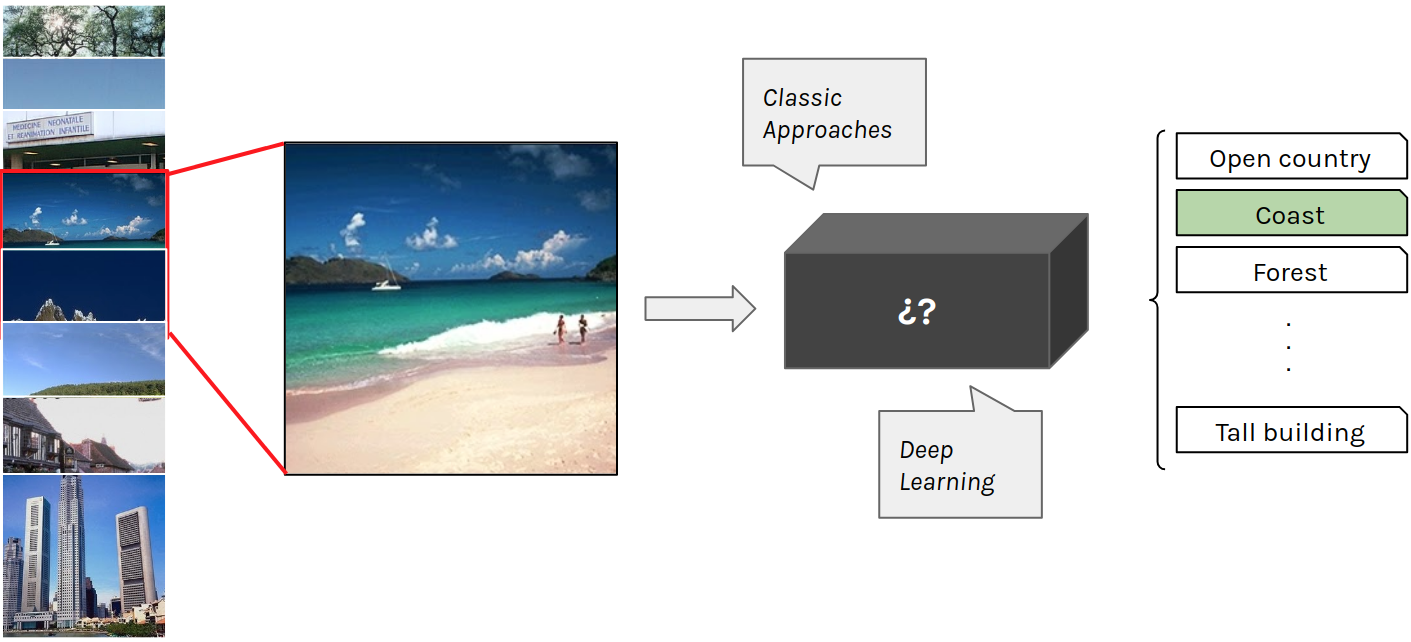}
    \caption{Simplified scheme of the image classification system} 
    \label{fig:image_classification}
\end{figure}

Given a dataset divided into 8 different classes, for each image in the dataset, the goal is to predict the class it belongs to. To do so, we implement and evaluate four different systems: BoVW approach, MLP based, and CNNs architectures: fine-tuning an existing one and designing one from scratch. In each case, the model is trained with a subset of images, and tested with unseen images to validate the performance by means of the accuracy and loss. In Fig.~\ref{fig:image_classification} we can see a simplified scheme of the system.

%% file: sections/02_data.tex
\section{DATA}
\label{sec:data}
\begin{figure}[t!]
\centering
\begin{tabular}{cccc}
\includegraphics[height=1.2cm]{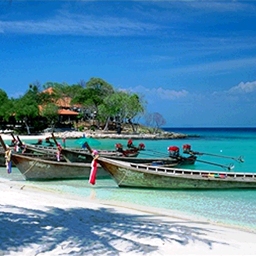} & 
\includegraphics[height=1.2cm]{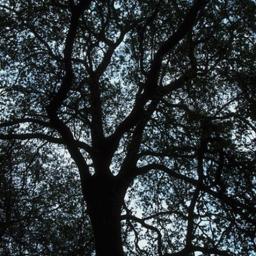} &
\includegraphics[height=1.2cm]{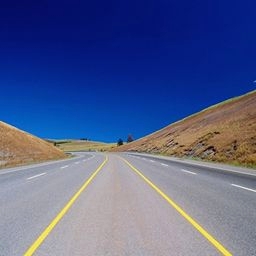} &
\includegraphics[height=1.2cm]{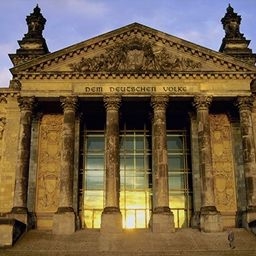} \\
(a) & (b) & (c) & (d) \\

\includegraphics[height=1.2cm]{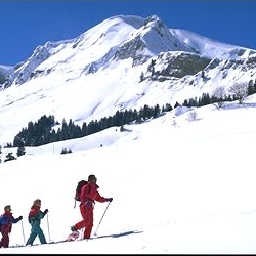} & 
\includegraphics[height=1.2cm]{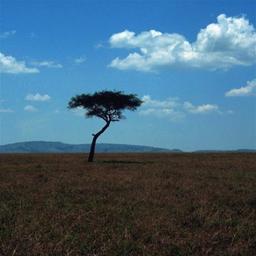} &
\includegraphics[height=1.2cm]{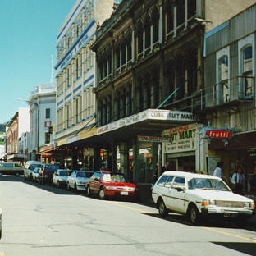} &
\includegraphics[height=1.2cm]{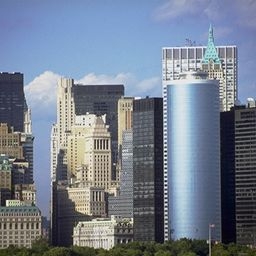} \\
(e) & (f) & (g) & (h) \\
\end{tabular}
\caption{Image sample from class (a) coast, (b) forest, (c) highway, (d) inside\_city, (e) mountain, (f) open\_country, (g) street and (h) tall\_building} \label{fig:sample_dataset}
\end{figure}

The dataset contains 2688 images from 8 different classes: coast, forest, highway, inside\_city, mountain, open\_country, street and tall\_building. In Fig.~\ref{fig:sample_dataset} a sample image from each class is presented. To properly train and evaluate the implemented systems, the dataset is divided into a training set of 1881 images (70\%) and a test set, that contains 807 images (30\%).

\begin{figure}[t!]
    \centering
    \includegraphics[width=0.9\linewidth]{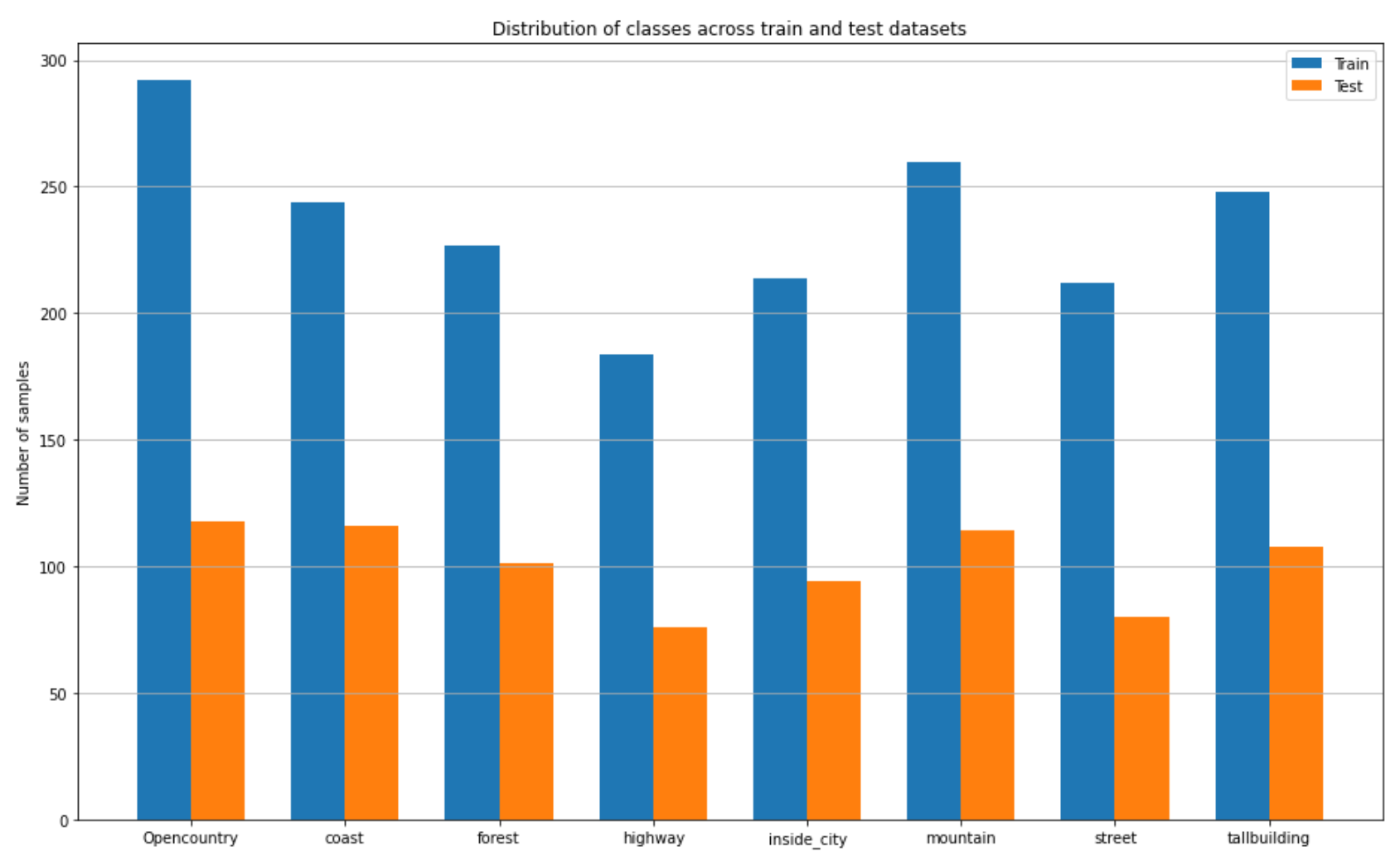} 
    \caption{Samples in each class for the train (blue) and test (orange) sets} \label{fig:balanced}
\end{figure}

Before start developing our system, it is important to analyze the dataset. For example, if the number of samples of each class is unevenly distributed (i.e. unbalanced dataset), using accuracy as the evaluation metric is not a good idea. In Fig.~\ref{fig:balanced} it can be observed that the data samples are more or less equally distributed across the different classes, so the dataset is balanced.

%% file: sections/03_method.tex
\input{sections/03_method/bovw}

\input{sections/03_method/mlp}

\input{sections/03_method/inception}

\input{sections/03_method/team4net}

\input{sections/03_method/aditya_part}

%% file: sections/03_method/bovw.tex
\section{Bag of Visual Words}
\label{sec:bovw}
\begin{figure}[t!]
\centering
\includegraphics[width=0.8\linewidth]{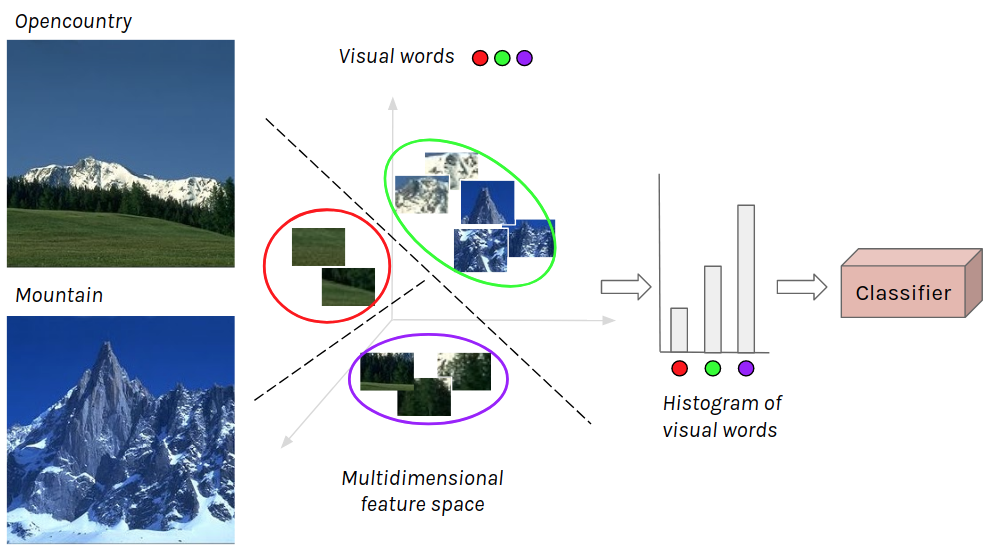} \\
\caption{Toy scheme of the BoVW classification system} 
\label{fig:bovw}
\end{figure}

The Bag of Visual Words (BoVW) approach consists on, given some training data, extract some local descriptors, cluster them in the multidimensional feature space to create visual words and count the number of words each image has (i.e. histogram of visual words). Therefore, a histogram is generated for each labeled image, and used to train a classifier such as Support Vector Machines (SVM). A toy scheme is presented in Fig.~\ref{fig:bovw}. In this section, the methods used to implement the BoVW system are explained in detail, and the results obtained with each configuration are presented and analyzed. For that purpose, we compute the accuracy with 8 (stratified) fold cross-validation in all cases.

\subsection{Keypoints and descriptors}
In the BoVW approach a feature detection algorithm is used to detect keypoints and extract local descriptors from each image, so the first step is to find which is the one that works best in our case using a k-Nearest Neighbors (KNN) classifier. The descriptors tested are: SIFT~\cite{sift} (vanilla and dense), SURF~\cite{surf} (vanilla and dense) and DAISY~\cite{daisy} (only dense).

\subsubsection{Vanilla descriptors}
In this scenario, keypoints are extracted using the detection algorithm of the corresponding local descriptor: SIFT or SURF (DAISY does not have any keypoint detector). An example of this detection is presented in Fig.~\ref{fig:keypoint_detection}b.

\subsubsection{Dense descriptors}
\begin{figure}[t!]
\centering
\begin{tabular}{ccc}
\includegraphics[height=1.7cm]{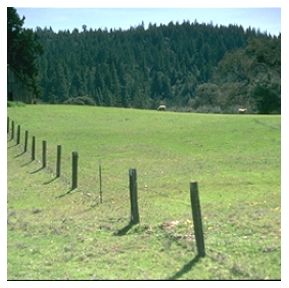} & 
\includegraphics[height=1.7cm]{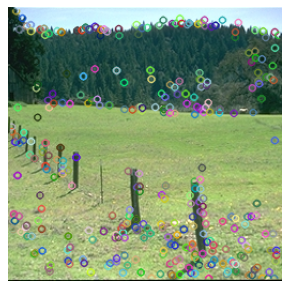} &
\includegraphics[height=1.7cm]{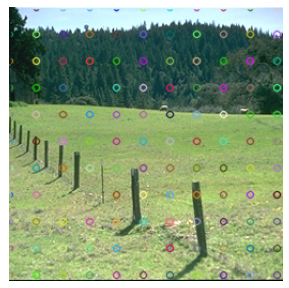} \\
(a) & (b) & (c) \\
\end{tabular}
\caption{(a) Image sample, (b) keypoint detection and (c) generated dense keypoints} \label{fig:keypoint_detection}
\end{figure}

Instead of using the detection algorithm to extract keypoints, we create a grid of spatially equidistant keypoints, which is translated into a more dense representation of the image. This is shown in Fig.~\ref{fig:keypoint_detection}c.

\input{tables/Bag_of_Visual_Words/bovw_descriptors}

The results are presented in Tab.~\ref{tab:bovw_descriptors}. As observed, the dense descriptors outperform vanilla SIFT and SURF. The main reason is that vanilla SIFT and SURF depend on the performance of the keypoint detector: if this detection is poor, the classification fails. When using dense keypoints, we get a representative descriptor of the images even if the content of the image is plain (low textures) or has repetitive patterns. In this case, the best results are obtained with dense SIFT, so when the \textbf{descriptor} is mentioned in this paper, it will be referring to \textbf{dense SIFT}.

The hyperparameters used to create the dense keypoints are also fine-tuned, and the results show that the best performance is obtained when the number of keypoints is larger. On the other hand, vanilla SIFT is scale-covariant, as it computes keypoints at different scales, so we tried to emulate this for dense SIFT with another parameter. However, there is not a substantial improvement in the results in this case, so the scale is not an important factor in this dataset. For this reason, we can conclude that losing the scale-covariance property of vanilla SIFT is not a problem when using dense SIFT.

\subsection{Classifiers}
The presented results are obtained using a k-NN classifier, which might be too simple in some tasks such as image classification. For this reason, we analyze the performance of the system using other classifiers: a logistic regression model and a SVM. As aforementioned, the descriptor used in each case is dense SIFT.

\input{tables/Bag_of_Visual_Words/bovw_classifiers}

The corresponding results, presented in Tab.~\ref{tab:bovw_classifiers}, show that SVM outperforms both k-NN and logistic regression classifiers. For this reason, we conclude that \textbf{SVM} is better suited for our problem, and thus it will be the one used in the rest of the experiments.

\subsubsection{Fine-tuning SVM kernel}
To further improve the results, we compute the accuracy (mean and standard deviation) for different SVM kernels. In addition to the the typical kernels (linear, poly, RBF and sigmoid), we create our own: the histogram intersection kernel, defined in Eq.~\ref{eq:histogram_kernel}.

\begin{equation}
    K_{i n t}(A, B)=\sum_{i=1}^{m} \min \left\{a_{i}, b_{i}\right\}
    \label{eq:histogram_kernel}
\end{equation}

The results are shown in Tab.~\ref{tab:bovw_svm_kernels}. The worse results are obtained using the linear kernel. The reason is that, in this dataset, images of different classes share visual words (e.g. trees in both forests and open country classes), so they are not linearly separable. On the other hand, the performance using the histogram intersection kernel is acceptable, as this kernel is useful in our specific problem, where the features are histograms. However, it is recommended to use the \textbf{RBF kernel}, which creates non-linear combinations of the features to uplift the samples onto a higher-dimensional feature space where a linear decision boundary can be used. Indeed, we obtain the best accuracy with this kernel.

\input{tables/Bag_of_Visual_Words/svm_kernel_w2}

\subsection{Spatial pyramids}
The BoVW system efficiently aggregates local features into a single global vector but it completely ignores the information about the spatial layout of the features. To tackle this, we compute the keypoints and descriptors at different pyramidal levels. Spatial pyramids work by partitioning the image into increasingly fine sub-regions and computing histograms of the visual words inside each sub-region.

\subsubsection{Square sub-regions}
The first approach is to divide each region in 4 square sub-regions for each level (Fig.~\ref{fig:bovw_spatial_pyramids}a,b). For example, using three levels, we first compute the descriptors for the whole image. Then, the image is divided in 4 blocks and the descriptors are computed for each of them. Each sub-block is divided into 4 blocks (16 in total) and the descriptors are computed for each of them. Finally, the descriptors computed in each level are concatenated (in this case, $1+4+16=21$ descriptors). By computing the descriptors of the different sub-regions of the image, we can later compute the histograms of visual words of each of the regions.

\subsubsection{Horizontal sub-regions}
\begin{figure}[t!]
\centering
\begin{tabular}{cc}
\includegraphics[height=1.5cm]{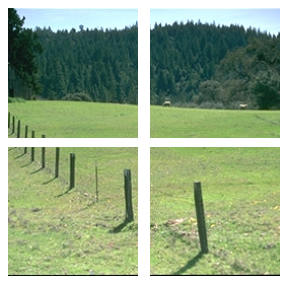} & 
\includegraphics[height=1.5cm]{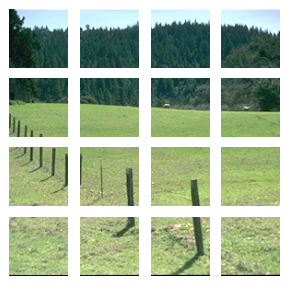} \\
(a) & (b)\\

\includegraphics[height=1.5cm]{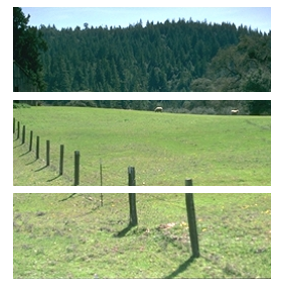} & 
\includegraphics[height=1.5cm]{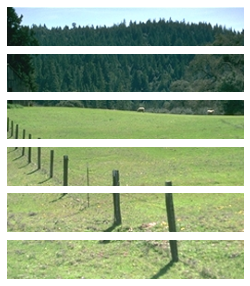} \\
(c) & (d)\\
\end{tabular}
\caption{Top: square sub-regions for levels (a) 1 and (b) 2. Bottom: horizontal sub-regions for levels 1 (c) and 2 (d)} 
\label{fig:bovw_spatial_pyramids}
\end{figure}

As this specific dataset is formed by a large number of landscape images, we thought it might be interesting to divide the image in horizontal sub-regions, as shown in Fig.~\ref{fig:bovw_spatial_pyramids}c,d. In this case, for each level the image is divided into 3 more sub-regions. For example, using three levels, the resulting descriptor will have ($1+3+6=$) 10 concatenated histograms.

\subsection{Normalization}
It is a good practice to normalize the data to avoid different scales of the feature vectors and thus improve data integrity. In this case, the histograms of visual words are normalized using L2 norm, Power norm or StandardScaler:

\begin{itemize}
    \item L2 norm, the modulus of the feature vector is 1:
    
\begin{equation}
    x_{norm}=\frac{x}{\|x\|_{2}}
    \label{eq:l2_norm}
\end{equation}

where 

\begin{equation}
    \|x\|_{2}=\sqrt{x_{1}^{2}+x_{2}^{2}+x_{3}^{2}+\ldots+x_{m}^{2}}
    \label{}
\end{equation}

    \item Power norm, the sum of all the values of the feature vector is 1:

\begin{equation}
    x_{norm}=\frac{x}{\sum_{i=1}(x_{i})}
    \label{eq:power_norm}
\end{equation}

    \item StandardScaler, standardize the feature vector by removing the mean and scaling to unit variance:

\begin{equation}
    x_{norm}=\frac{x-\mu}{\sigma}
    \label{eq:stdscaler}
\end{equation}

where $\mu$ is the mean and $\sigma$ the standard deviation.

\end{itemize}

The normalization of the histograms of visual words play an important role in the spatial pyramid algorithm. If the image is divided in 4 blocks, each block has $1/4$ of the information of the whole image. Therefore, when computing the histogram of a block, the energy (and thus the contribution) of that histogram will be $1/4$ of the energy of the histogram of the whole image. To give the same importance to each of them, we normalize all histograms so that they contribute the same.

The results obtained using square and horizontal sub-regions and without and with normalization are presented in Tab.~\ref{tab:square_pyramid_w2}. As observed, for this specific dataset, if the image is divided in \textbf{horizontal sub-regions}, the results are better. The reason is that most of the images are landscape images with easily differentiated horizons, as aforementioned. Comparing the results for the different pyramid levels, it is observed that levels 1 and 2 outperform level 0 in all cases, which proves that features \textbf{spatial information is relevant} to image classification performance. On the other hand, level 2 and level 1 only differ by a small margin in most of the cases, but level 2 comes with a big extra computational cost (e.g. 21 vs 5 histograms for the square sub-regions), so \textbf{level 1} is chosen as the most suitable pyramidal level to use. For higher pyramidal levels, curse of dimensionality also comes into picture as our feature space becomes sparse. Regarding normalization, we observe that the results are slightly improved in some cases, specially for the L2 norm and standardScaler. However, the improvement is not significant, which validates the consistency and integrity of our data. Even so, it is a good practice to normalize the data, and it will be useful to improve our results later, so \textbf{StandardScaler normalization} will be used.

\input{tables/Bag_of_Visual_Words/square_pyramid_w2}

\subsection{Clustering}
\begin{figure}[t!]
\centering
\includegraphics[width=0.7\linewidth]{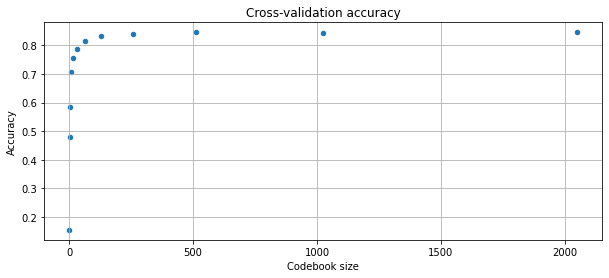} \\
\caption{Cross-validation accuracy for different codebook sizes} 
\label{fig:bovw_codebook_size}
\end{figure}

To create the visual words, the multidimensional feature space is clustered using k-means, being k the codebook size (number of visual words). To further improve the results, this hyperparameter is fine-tuned, and the results are presented in Fig.~\ref{fig:bovw_codebook_size}. As observed, with a small codebook size (e.g. 32), the visual words are too general, so they are not representative enough to perform classification properly. With larger codebook sizes, the results improve up to a certain point (0.85 for a codebook size of 512), as with more visual words each class is well represented, so it is easier for the classifier to distinct between them. The best results are obtained with a \textbf{codebook size of 512}, so that will be the one used. With codebook sizes of 1024 and 2048 the results are also good, but the computational cost is much higher.

\subsection{Reducing dimensionality}

The best results are obtained with large codebook sizes, such as 512, so we use Principal Component Analysis (PCA) to decrease the computational time. Concretely, PCA is used to reduce the feature space dimensionality projecting it to a lower dimensional space. This reduction of dimensionality is very useful in our case, as spatial pyramids increase the dimension of the vector and thus the computational time. Gridsearch is performed to fine-tune the parameter \textit{num\_components}, which is used to select the number of dimensions to be kept after the dimensionality reduction. 

\begin{figure}[t!]
\centering
\includegraphics[width=0.7\linewidth]{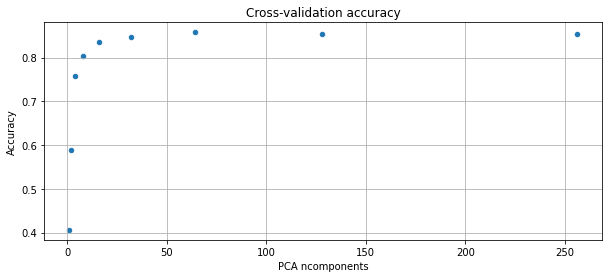} \\
\caption{Cross-validation accuracy for different PCA num\_components} 
\label{fig:bovw_pca}
\end{figure}

As shown in Fig.~\ref{fig:bovw_pca}, the best results are obtained with the larger \textit{num\_components} (up to 0.87). This parameter defines the dimensionality of the resulting vector, and the higher its dimension, the more representative of the data and thus the better the performance of the classifier. Applying PCA slightly improves the performance, but more importantly, it speeds up the computation. For this reason, \textbf{PCA with \textit{num\_components=64}} will be used.

\subsection{Fisher vectors}

Even if the BoVW approach performs well on our dataset, it finds the closest word in the vocabulary relying only on the number of local descriptors assigned to each Voronoi cell. With fisher vectors, we are not only using the mean of the local descriptors, but also including higher order statistics: the covariance. This way, the information of how far is each feature from its closest vocabulary word (and also to the other vocabulary words) is obtained.

To study its performance in our dataset, we fit an SVM classifier with the train fisher vectors (obtained from the training dataset) and we use the test fisher vectors to predict the labels of the test dataset. Fisher vectors allows training more discriminative classifiers with a lower vocabulary size. The obtained results show that encoding our feature vectors using second order information (covariances along with means) indeed benefits classification performance, as it provides similar results (0.84) reducing the computational cost.

%% file: tables/Bag_of_Visual_Words/bovw_descriptors.tex
\begin{table}[b!]
\centering
\caption{Accuracy obtained with different descriptors using k-nn}
\label{tab:bovw_descriptors}

\renewcommand{\arraystretch}{1.3}
\setlength\tabcolsep{2pt}

\begin{tabular}{{c|cc}}
\toprule
    Descriptor & Type & Accuracy \\ 
    \midrule
    \midrule
    \multirow{2}{*}{\textbf{SIFT}} & Vanilla & 0.55 \\
        {} & \textbf{Dense} & \textbf{0.74}  \\
    \hline
    \multirow{2}{*}{SURF} & Vanilla & 0.62 \\
        {} & Dense & 0.63 \\
    \hline
    DAISY & Dense & 0.66\\
\bottomrule
\end{tabular}
\end{table}

%% file: tables/Bag_of_Visual_Words/bovw_classifiers.tex
\begin{table}[t!]
\centering
\caption{Accuracy obtained with different classifiers using dense SIFT}
\label{tab:bovw_classifiers}

\renewcommand{\arraystretch}{1.3}
\setlength\tabcolsep{2pt}

\begin{tabular}{{c|c}}
\toprule
    Classifier & Accuracy \\ 
    \midrule
    \midrule
    k-NN & 0.74\\
    \hline
    Logistic Regression & 0.79\\
    \hline
    SVM & 0.83\\
\bottomrule
\end{tabular}
\end{table}

%% file: tables/Bag_of_Visual_Words/svm_kernel_w2.tex
\begin{table}[b!]
\centering
\caption{Accuracy obtained with different SVM kernels}
\label{tab:bovw_svm_kernels}
\begin{tabular}{{c|cc}}
\toprule
\multirow{2}{*}{Kernel} & \multicolumn{2}{c}{Accuracy}\\
{} & Mean & Std Dev\\ 
\midrule
\midrule
Linear & 0.77 & 0.03\\
Poly & 0.76 & 0.03 \\
\textbf{RBF} & \textbf{0.83} & \textbf{0.02}\\
Sigmoid & 0.81 & 0.02 \\
Histogram intersection & 0.81 & 0.02\\
\bottomrule
\end{tabular}
\end{table}

%% file: tables/Bag_of_Visual_Words/square_pyramid_w2.tex
\begin{table}[t!]
\centering
\caption{Accuracy obtained combining different pyramid shapes, levels and normalizations}
\label{tab:square_pyramid_w2}

\renewcommand{\arraystretch}{1.3}
\setlength\tabcolsep{2pt}

\begin{tabular}{{c|c|cccc}}
\toprule
\multirow{2}{*}{Shape} & \multirow{2}{*}{Level} & \multicolumn{4}{c}{Normalization} \\
        {} & {} & None & L2 & Power & \textbf{StandardScaler} \\ 
\midrule
\midrule
\multirow{3}{*}{Square} & 0 & 0.81 & 0.81 & 0.81 & 0.81\\
{} & 1 & 0.81 & 0.82 & 0.81 & 0.82\\
{} & 2 & 0.82 & 0.83 & 0.82 & 0.83\\
\hline
\multirow{3}{*}{\textbf{Horizontal}} & 0 & 0.81 & 0.81 & 0.81 & 0.81\\
{} & \textbf{1} & 0.82 & 0.83 & 0.83 & \textbf{0.84}\\
{} & 2 & 0.82 & 0.83 & 0.83 & 0.83\\
\bottomrule
\end{tabular}
\end{table}

%% file: sections/03_method/mlp.tex
\section{MLP}
\label{sec:mlp}
The results obtained with a classic approach such as the Bag Of Visual Words system are acceptable, but not good enough to consider the implemented image classifier robust nor reliable. For this reason, we need to use advanced techniques to improve the performance and obtain the desired results: the well-known Deep Learning (DL). As a first step towards DL, we explore the most simple architecture: Multilayer Perceptrons (MLP), in which each neuron of each layer is connected (forwards).

\begin{figure}[t!]
\centering
\begin{tabular}{cc}
\includegraphics[height=2.7cm]{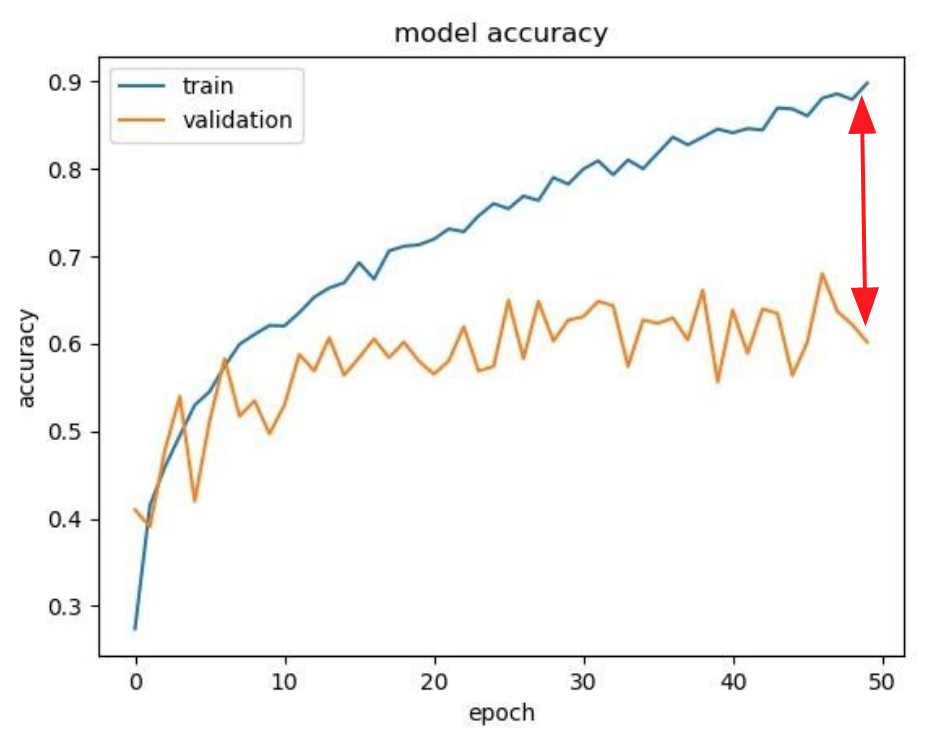} & 
\includegraphics[height=2.7cm]{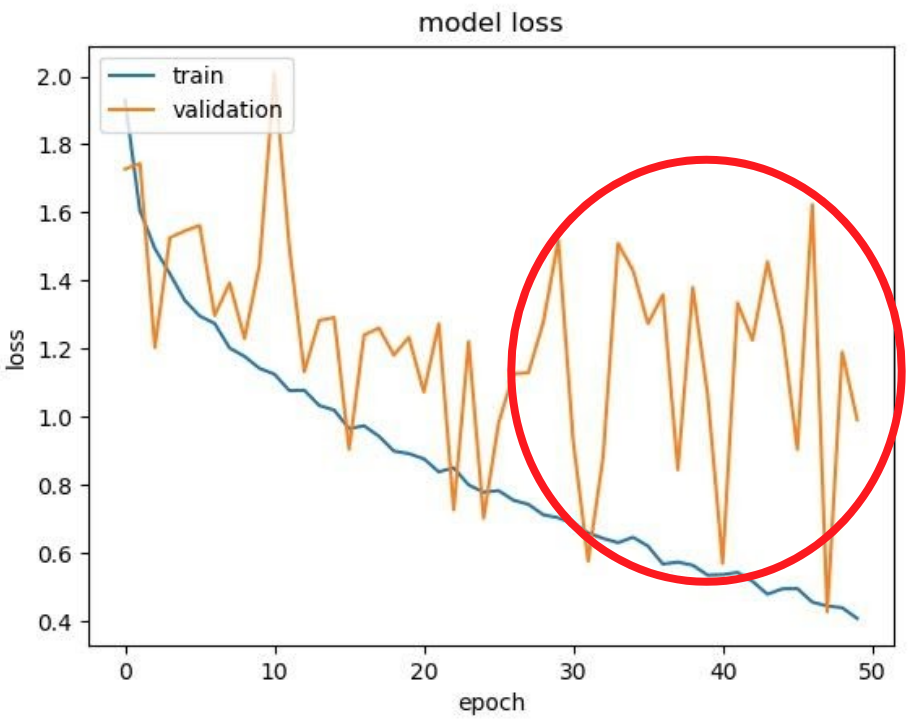} \\
(a) & (b) \\
\end{tabular}
\caption{MLP: training and validation (a) accuracy and (b) loss curves} \label{fig:mlp_accuracy_loss}
\end{figure}

Using a simple MLP and a \textit{softmax} layer (last layer used to predict the class of each input image), the results in terms of accuracy and loss are really bad, as shown in Fig.~\ref{fig:mlp_accuracy_loss}. The difference between the train and validation accuracy curves is an indicator that the model is overfitting to the training data, and thus is not able to generalize well to unseen samples (those of the test data). Moreover, the validation loss curve is unstable and not properly minimized.

To try to obtain better results, we fine-tune different parameters: learning rate, image size, number of layers (depth), layers sizes, adding normalization or regularization, and so on. Even if the performance is slightly improved in some cases, the potential of the system is limited to the fact that we are using a simple MLP for a hard image classification task, so the results will not get better. It is worth mentioning that the impact of each of these parameters (e.g. learning rate) is deeply studied in Section ~\ref{sec:own_cnn}, where we design a CNN from scratch, which is more realistic in the image classification task.

\subsection{Deep Features, SVM and BoVW}
Finally, before moving on to CNNs, we explore different variations of the MLP system in order to see if the results can be improved: 

\begin{enumerate}
    \item \textbf{Deep Features (DF) + SVM}: Extract DF from the deeper hidden layer (previous to \textit{softmax}, where the features are more abstract/general) and use them to train an SVM classifier.
    \item \textbf{Aggregating predictions}: Divide the input image into small patches, extract the prediction for each patch and aggregate the final prediction.
    \item \textbf{DF as a dense descriptor + BoVW} Divide the input image into small patches, extract the DF from the last hidden layer for each patch, and concatenate them to create a feature vector for each image. Then, use k-means to create a codebook and train an SVM classifier with the histograms of visual words.
\end{enumerate}

The obtained results are presented in Tab.~\ref{tab:mlp_variations_df_svm}. As observed, extracting DF and using them to train an SVM classifier is not a good alternative. Another approach is to divide each image in patches and extract deep features from each of them. In this other two cases, even if the results are slightly improved, they are not acceptable, and much worse than the ones obtained with the classic BoVW approach. For this reason, we conclude that MLP is too simple for this image classification problem.

\input{tables/MLP/accuracy_mlp}

%% file: tables/MLP/accuracy_mlp.tex
\begin{table}[t!]
\centering
\caption{Accuracy obtained using the baseline and the different improvements on MLP}
\label{tab:mlp_variations_df_svm}

\renewcommand{\arraystretch}{1.3}
\setlength\tabcolsep{2pt}

\begin{tabular}{{c|c}}
\toprule
MLP Configuration & Accuracy\\ 
\midrule
\midrule
Baseline & 0.61\\
Optimal Learning Rate & 0.61 \\
Different Image Sizes & 0.64 \\
Different Layer Sizes & 0.65\\
Increasing Depth & 0.66 \\
DF + SVM & 0.53 \\
Aggregating predictions & 0.72\\
DF as a dense descriptor + BoVW & 0.70 \\
\bottomrule
\end{tabular}
\end{table}

%% file: sections/03_method/inception.tex
\section{InceptionV3}
\label{sec:inception}
The results obtained with MLP are not acceptable, so we take a step forward into deep learning to use the state of the art architecture in image-related tasks: Convolutional Neural Networks (CNNs). Since the outbreak of CNN in 2012 with AlexNet~\cite{AlexNet}, multiple architectures have been presented to tackle the classification problem, obtaining increasingly better results in terms of minimizing the miss-classification error. In this paper, we fine-tune InceptionV3~\cite{inception} to adapt it to our specific dataset.  This network, created by Google, is based on the idea of using Inception modules to use different sizes of channels in parallel, as there are four parallel channels in each module, which are further concatenated at the end. Specifically, each module includes factorizing convolutions with large filter size into smaller filter, factorization into asymmetric convolutions and auxiliary classifiers introduced to tackle the problem of vanishing gradient.

This model has been trained and tested with the ImageNet dataset~\cite{imagenet}, which contains around 1M images divided in 1000 classes. Therefore, it has not much to do with the dataset used in this study, and we need to adapt InceptionV3 to our specific problem.

\subsection{Changing the network architecture}
The first approach we take is to use the existing model and weights by modifying only the last layer of the architecture: the \textit{softmax} layer. This is a required step to adapt the output to the number of classes that our dataset has: eight. First of all, we freeze all the layers except the last one, so that the training stage does not affect the pre-trained weights of the model.

\begin{figure}[t!]
\centering
\begin{tabular}{cc}
\includegraphics[height=2.7cm]{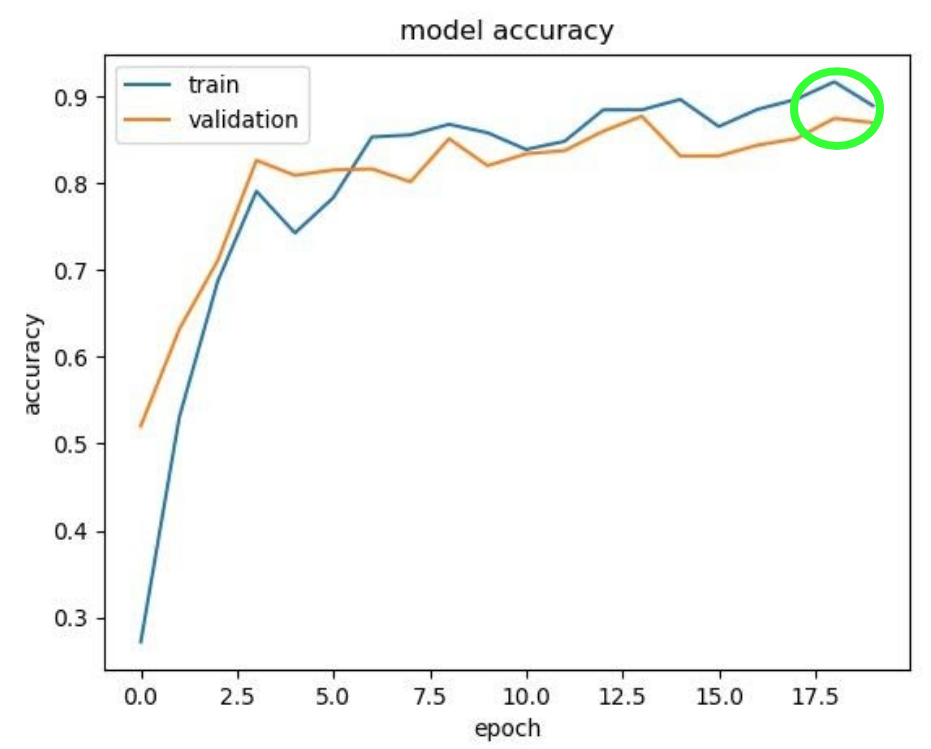} & 
\includegraphics[height=2.7cm]{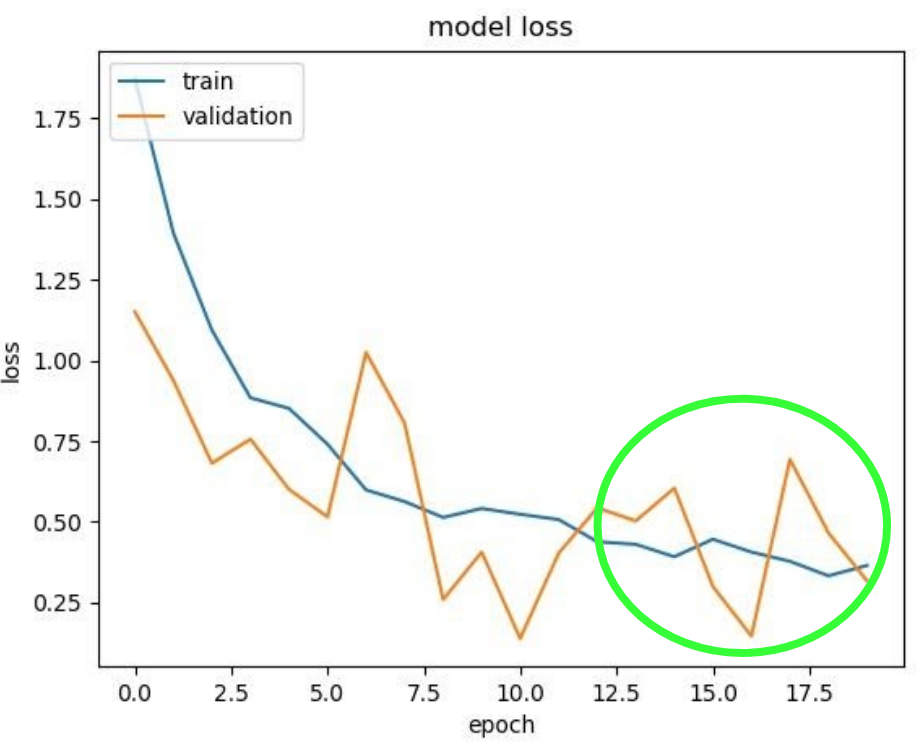} \\
(a) & (b) \\
\end{tabular}
\caption{InceptionV3: training and validation (a) accuracy and (b) loss curves} \label{fig:inception_accuracy_loss}
\end{figure}

As observed, in Fig.~\ref{fig:inception_accuracy_loss}, the results are greatly improved using InceptionV3 compared to the ones obtained with a simple MLP, which proves the potential and usefulness of CNNs in image classification problems. Specifically, the difference between the training and validation loss is much lower, so there is no overfitting. Furthermore, both training and validation losses are correctly minimized.

\begin{figure}[t!]
\centering
\begin{tabular}{cc}
\includegraphics[height=2.9cm]{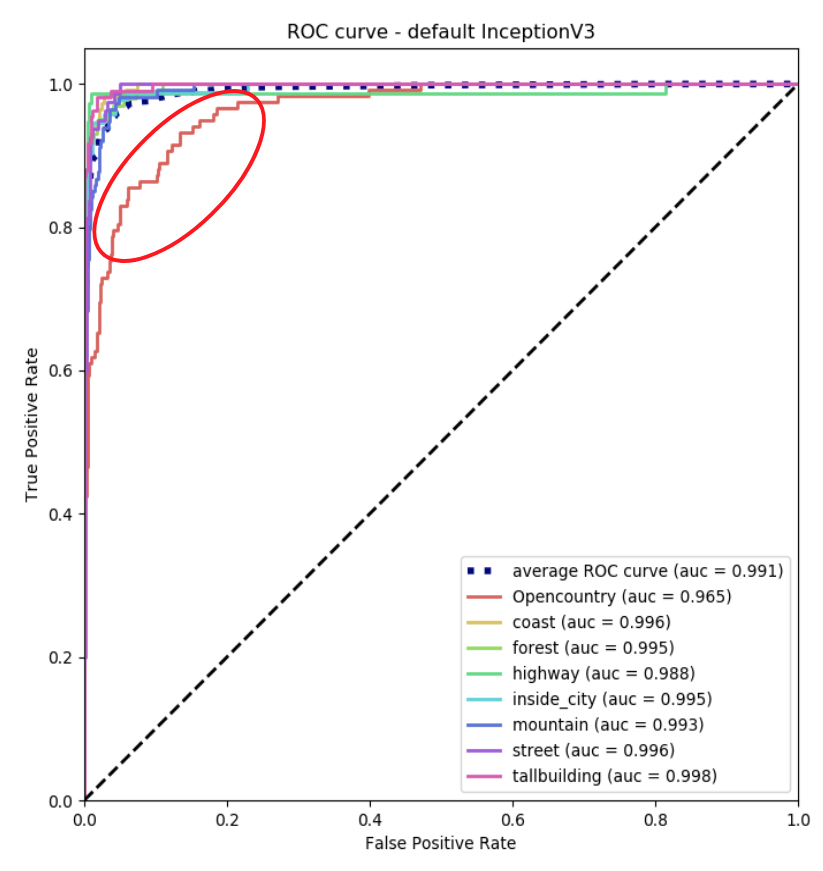} & 
\includegraphics[height=2.9cm]{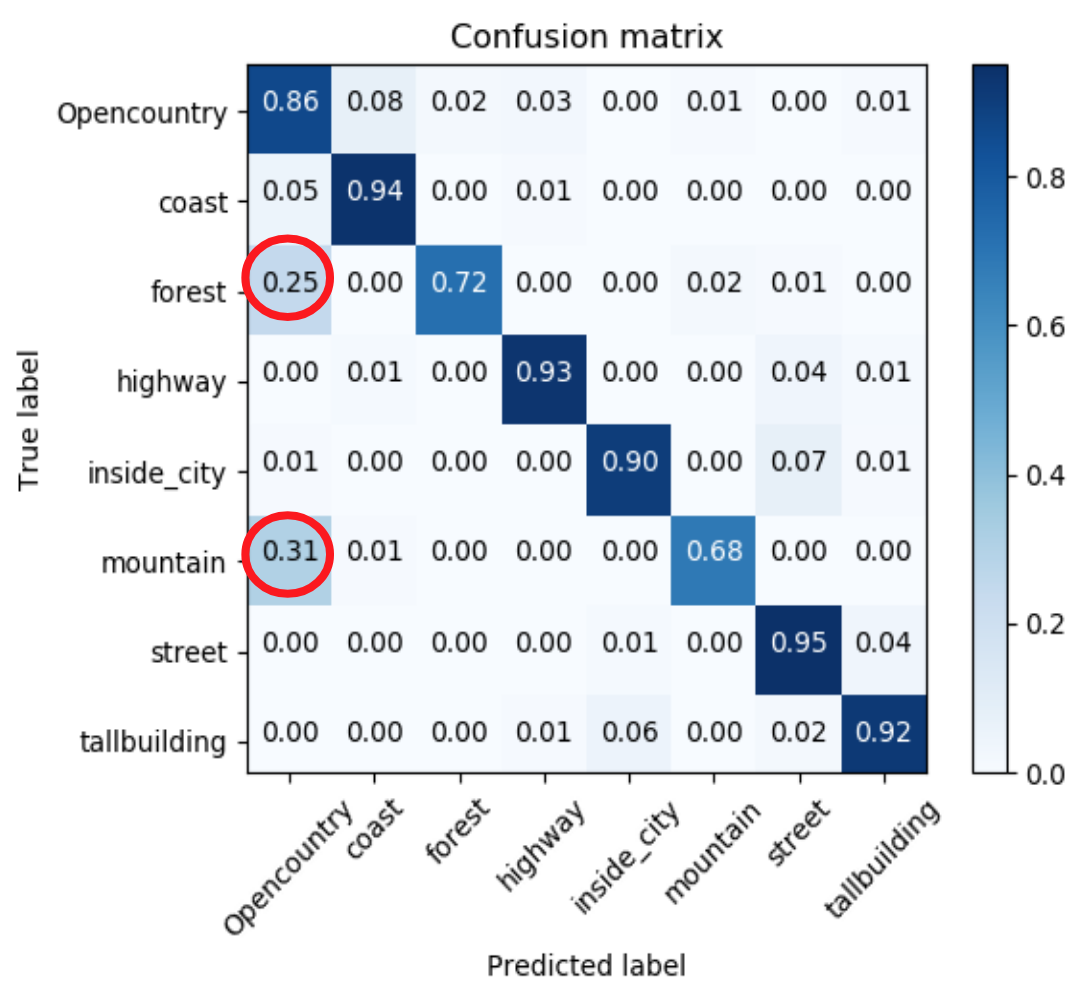} \\
(a) & (b) \\
\end{tabular}
\caption{InceptionV3: (a) ROC curves and (b) confusion matrix} 
\label{fig:inception_roc_confusion}
\end{figure}

In the confusion matrix (Fig.~\ref{fig:inception_roc_confusion}b), we observe that InceptionV3 performs really well in most of the cases, but it misclassifies a lot of forest and mountain samples as opencountry. This is also observed in the ROC curve (Fig.~\ref{fig:inception_roc_confusion}a), as the opencountry’s Area Under the Curve (AUC) is lower.

\subsection{Unfreezing some layers}
The next step is to unfreeze and retrain some layers of InceptionV3, to see if the learned weights improve the results. InceptionV3 is divided in 11 Inception blocks and a total of 311 layers, so it is not easy to select which layers to unfreeze. For this reason, we unfreeze by blocks (always starting from the end).

\input{tables/inceptionv3/inception_unfreezing}

The results are presented in Tab.~\ref{tab:inception_unfreezing}. As expected, the test accuracy increases as we unfreeze and retrain more blocks of the model, being the best result the one obtained with the full retrained model. However, the number of trainable parameters also increases, so the computational cost is much higher.

\subsection{Removing blocks of layers from InceptionV3}
In order to reduce the number of parameters, we decide to remove some Inception blocks and study the performance of the new (retrained) model in our specific dataset. To do so, we get the output from an specific block (e.g. block \#3, so that 8 blocks are removed), add a \textit{globalaveragepooling2d} layer and a final \textit{softmax} layer. In each case, the full model is retrained and it takes a different number of epochs to converge.

\input{tables/inceptionv3/inception_removing}

As observed in Tab.~\ref{tab:inception_removing}, by removing 5 blocks we still obtain a really high accuracy, and we reduce the number of parameters from 21.82M to 5M, more than four times less, which makes our system lighter. This is possible because our dataset is much simpler than ImageNet, as it only has 8 different classes, so the model needs a much smaller number of parameters to learn to predict them.

\subsection{Tiny dataset}
\begin{figure}[t!]
\centering
\begin{tabular}{cc}
\includegraphics[height=2.7cm]{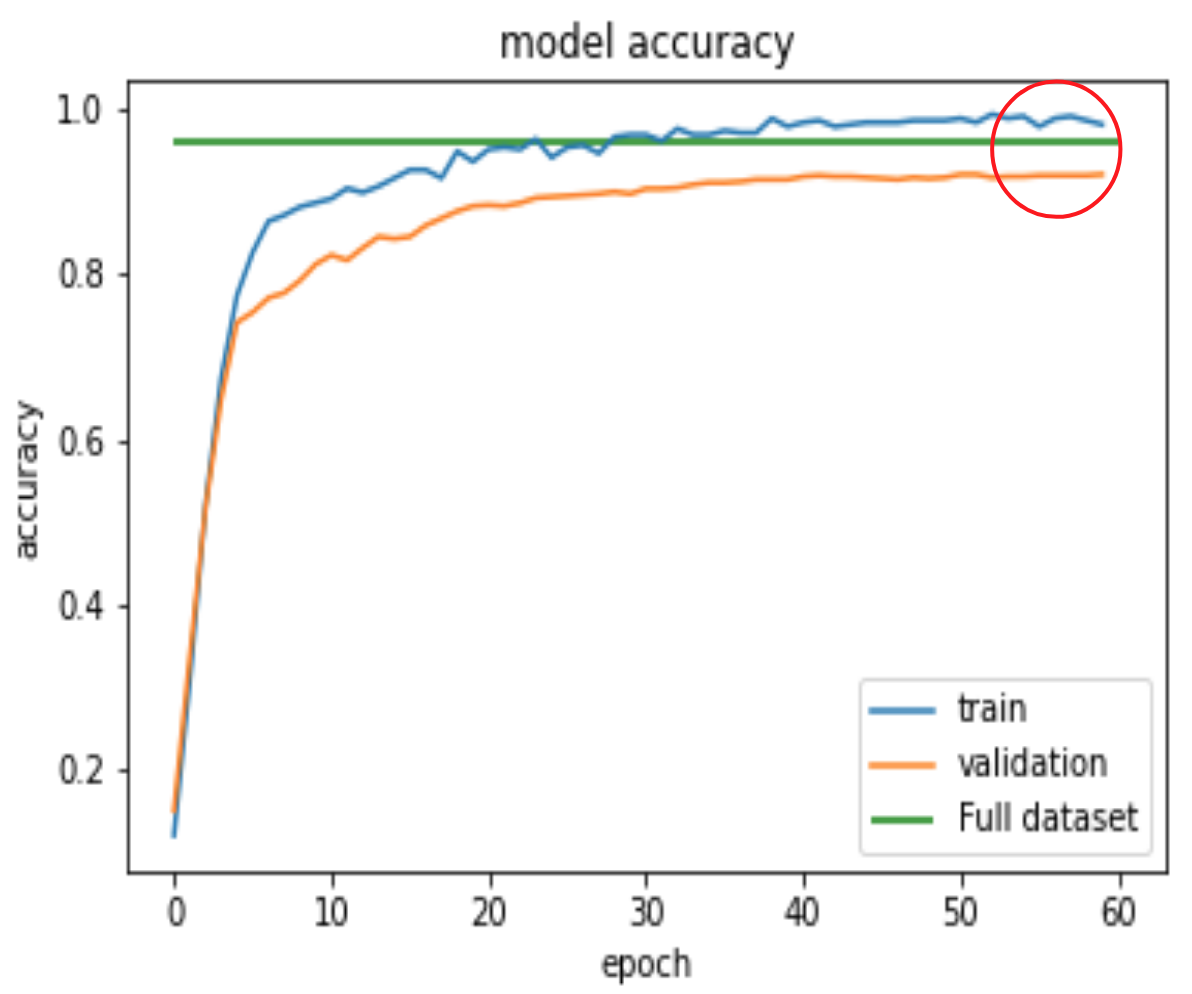} & 
\includegraphics[height=2.7cm]{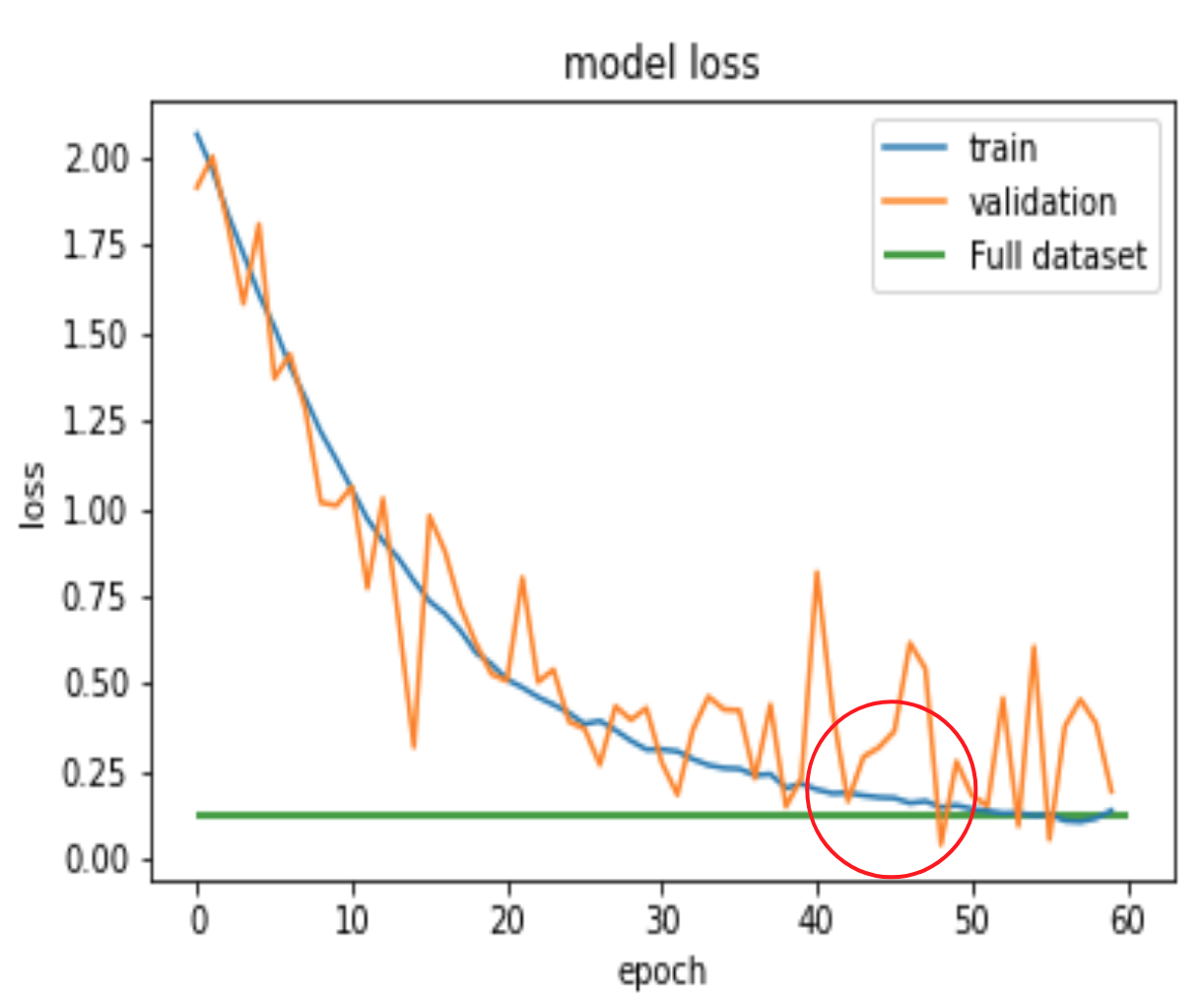} \\
(a) & (b) \\
\end{tabular}
\caption{New model trained with tiny dataset: training and validation (a) accuracy and (b) loss curves} 
\label{fig:inception_tiny_accuracy_loss}
\end{figure}

Once the architecture of our model has been lightened, we train it with a smaller dataset (50 samples per class, making a total of 400 samples) to study its performance. This is presented in Fig.~\ref{fig:inception_tiny_accuracy_loss}. As expected, the model is not able to learn that good (nor fast) using the tiny dataset, as it needs more samples to correctly set the weights of each layer, and the resulting accuracy is lower.

To improve the learning of our new model with the tiny dataset, we introduce and evaluate the usage of data augmentation. To do so, we use different augmentations individually and combined, to see if adding more variability to our training data improves the performance of the new model.

\input{tables/inceptionv3/augmentations}

The results, shown in Tab.~\ref{tab:inception_augmentations}, show that each one of the augmentation methods is helping our model, so data augmentation is very useful to contribute to the variability of the training data and thus help with the learning. However, combining the augmentations the results are not further improved, as they may distort the images too much.  Therefore, horizontal flip is enough for this problem.

\subsection{Random search to tune the hyperparameters}
The last step is to refine the model hyperparameters that optimizes the validation accuracy results using the following options:

\begin{itemize}
    \item Optimizer: SGD, RMSprop, Adam, Adadelta, Adagrad
    \item Learning Rate: 0.001, 0.01, 0.1, 0.2
    \item Momentum: 0.6, 0.8, 0.9
    \item Activation function: elu, relu, tanh
\end{itemize}

Considering the size of our network, we cannot do an exhaustive gridsearch, as it is not feasible in terms of computational time, so we use the random search implementation from keras tuner.

\begin{figure}[b!]
\centering
\begin{tabular}{cc}
\includegraphics[height=2.7cm]{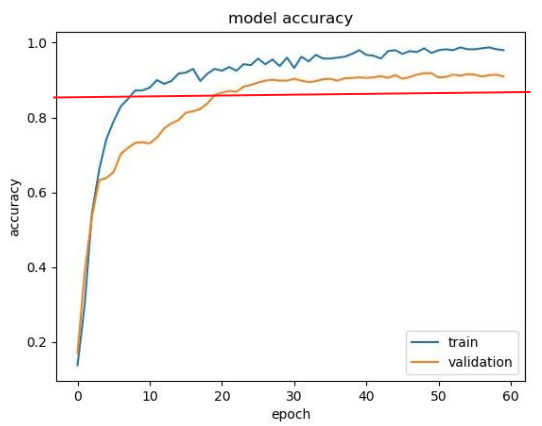} & 
\includegraphics[height=2.7cm]{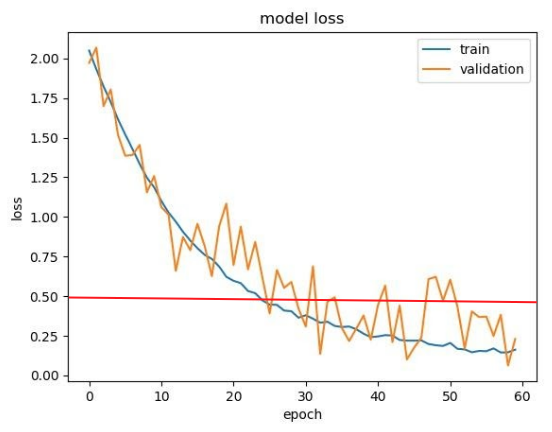} \\
(a) & (b) \\
\end{tabular}
\caption{Final model trained with tiny dataset: training and validation (a) accuracy and (b) loss curves. Red line: original InceptionV3 results} 
\label{fig:inception_tiny_final_accuracy_loss}
\end{figure}

The best results are obtained with the SGD optimizer, learning rate of 0.001, momentum of 0.9 and relu activation function. The resulting accuracy and loss curves are presented in Fig.~\ref{fig:inception_tiny_final_accuracy_loss}. As expected, our model needs more epochs to converge compared to the original InceptionV3, as now we are using a much smaller dataset. However, the resulting accuracy is higher, and the loss is correctly minimized, which proves that the network has been correctly fine-tuned for our specific case.

\begin{figure}[t!]
\centering
\begin{tabular}{cc}
\includegraphics[height=2.7cm]{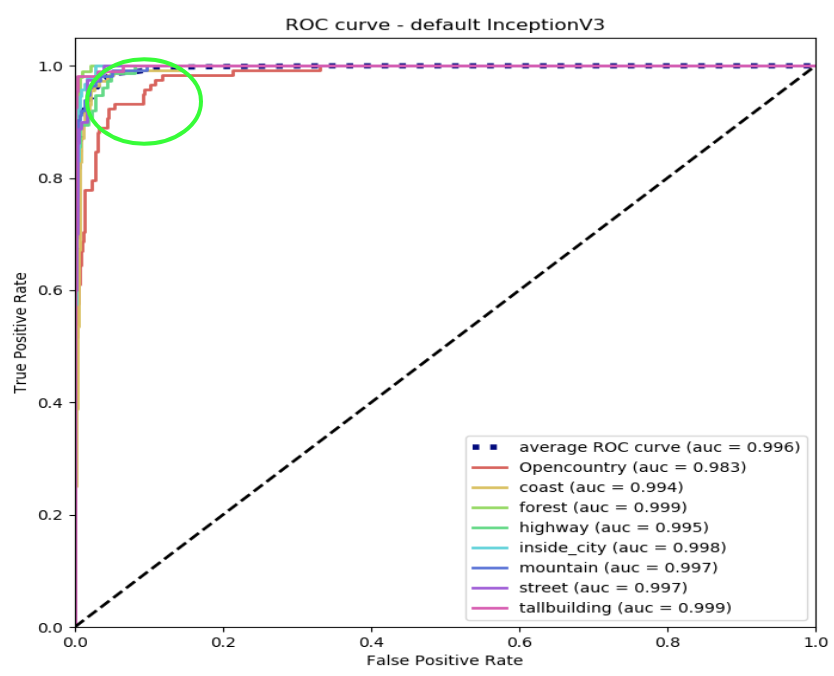} & 
\includegraphics[height=2.7cm]{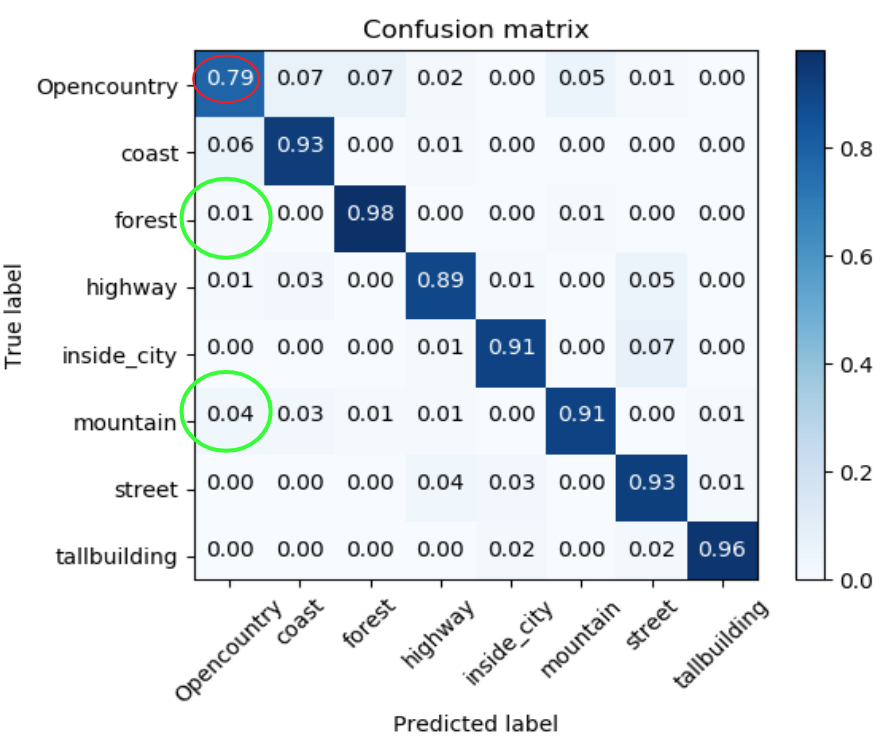} \\
(a) & (b) \\
\end{tabular}
\caption{Final model trained with tiny dataset: (a) ROC curves and (b) confusion matrix} 
\label{fig:inception_tiny_final_roc_confusion}
\end{figure}

The ROC curve and confusion matrix are also presented in Fig.~\ref{fig:inception_tiny_final_roc_confusion}. As observed, the new model has improved the performance in the classes that were more difficult for the original InceptionV3, as it is not misclassifying forest and mountain samples as opencountry anymore. However, the recall of the opencountry class has decreased, so this model could be further improved.

%% file: tables/inceptionv3/inception_unfreezing.tex
\begin{table}[t!]
\centering
\caption{Accuracy and parameters for different number of unfreezed Inception blocks}
\label{tab:inception_unfreezing}

\renewcommand{\arraystretch}{1.3}
\setlength\tabcolsep{2pt}

\begin{tabular}{{c|ccc}}
\toprule
\multirow{2}{*}{Unfreezed blocks} & \multicolumn{2}{c}{Parameters (Total: 21.82M)} & \multirow{2}{*}{Accuracy}\\
{} & Trainable & Non-trainable & {}\\ 
\midrule
\midrule
None & 16.3K & 21.81M & 0.89 \\
1 & 6.09M & 15.73M & 0.91 \\
3 & 12.83M & 8.99M & 0.92 \\
5 & 16.66M & 5.16M & 0.94 \\
7 & 19.64M & 2.18M & 0.93 \\
\textbf{All} & 21.78M & 34.43K & \textbf{0.96} \\
\bottomrule
\end{tabular}
\end{table}

%% file: tables/inceptionv3/inception_removing.tex
\begin{table}[t!]
\centering
\caption{Accuracy and parameters for different number of removed Inception blocks}
\label{tab:inception_removing}

\renewcommand{\arraystretch}{1.3}
\setlength\tabcolsep{2pt}

\begin{tabular}{{c|ccc}}
\toprule
Removed blocks & Parameters & Epochs & Accuracy\\
\midrule
\midrule
None & 21.82M & 20 & 0.96 \\
1 & 15.7 & 40 & 0.96 \\
3 & 8.9M & 50 & 0.94 \\
\textbf{5} & \textbf{5M} & 60 & \textbf{0.96} \\
7 & 2.1M & 70 & 0.90 \\
\bottomrule
\end{tabular}
\end{table}

%% file: tables/inceptionv3/augmentations.tex
\begin{table}[t!]
\centering
\caption{Accuracy for different data augmentations in the tiny dataset}
\label{tab:inception_augmentations}

\renewcommand{\arraystretch}{1.3}
\setlength\tabcolsep{2pt}

\begin{tabular}{{c|cc}}
\toprule
Data augmentation & Value & Accuracy\\
\midrule
\midrule
None & - & 0.90 \\
Horizontal Flip (HF) & True & 0.93 \\
Zoom (Z) & 20\% & 0.92 \\
Rotate (R) & 10º & 0.92 \\
Shear (S) & 20\% & 0.92 \\
Width Shift (WS) & 20\% & 0.92 \\
Height Shift (HS) & 20\% & 0.92 \\
\bottomrule
\end{tabular}
\end{table}

%% file: sections/03_method/team4net.tex
\section{Designing our own CNN}
\label{sec:own_cnn}
\begin{figure}[t!]
    \centering
    \includegraphics[height=0.3\textwidth]{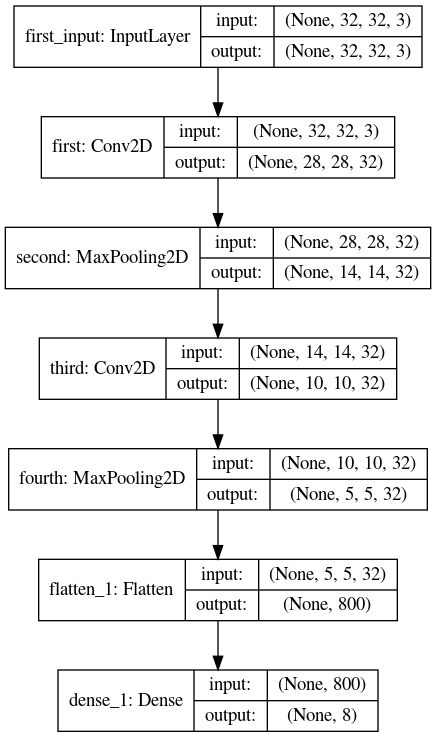} 
    \caption{Baseline architecture} \label{fig:baseline_model}
\end{figure}

To better fit the model to the problem, we design a CNN from scratch. The baseline of our network is formed by two blocks of a 2D convolutional layer and a 2D max pooling, followed by a dense output layer with a 'softmax' activation function. The model is represented in Fig.~\ref{fig:baseline_model}.

\begin{figure}[t!]
\centering
\begin{tabular}{cc}
\includegraphics[width=0.35\linewidth]{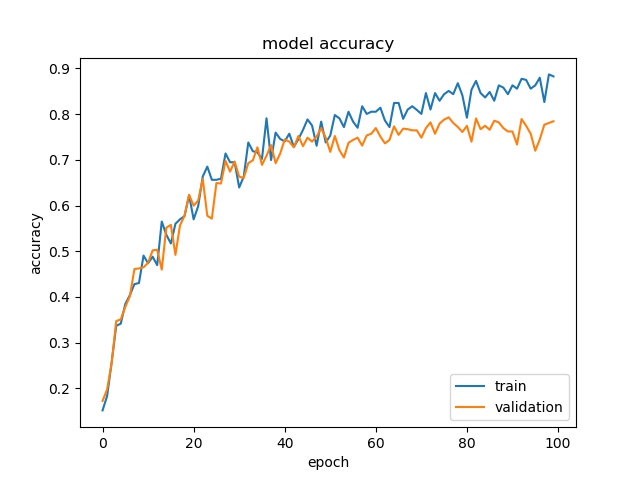} &
\includegraphics[width=0.35\linewidth]{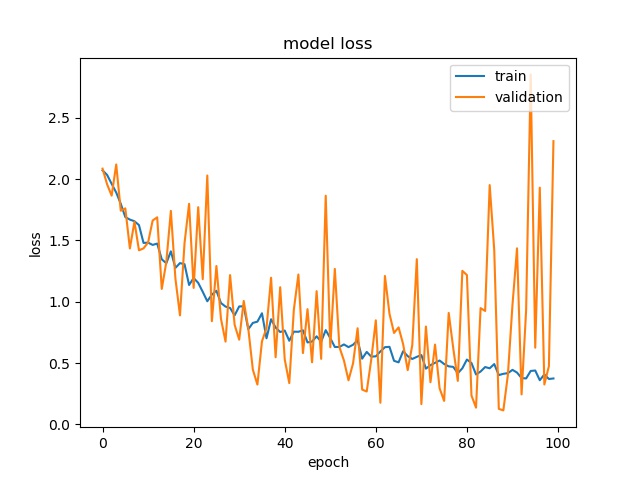} \\
(a) & (b)
\end{tabular}
\caption{ Baseline (a) accuracy and (b) loss curves} \label{fig:baseline_metrics}
\end{figure}

The performance of this baseline model can be seen in terms of accuracy and loss in Fig.~\ref{fig:baseline_metrics}.

We obtain an accuracy of 0.78, which is already more than what we get with the MLP. However, the accuracy curve shows overfitting and the loss curve is unstable and starts diverging for the validation set.

The parameters used on the convolutional layers are the default ones by Keras examples: Kernel size of 5x5, 32 filters, Relu activation function and Glorot Normal weight initialization.

\subsection{Kernel Size}
In order to improve our system, we tune the different parameters of the convolutional layers to find out the limits of this baseline. 
The first parameter to be tuned is the kernel size, obtaining the results in Tab.~\ref{tab:kernel_tuning}. The best accuracy is obtained with the kernel sizes of 5x5 and 7x7. However, for our CNN we introduce another metric to take into account: the accuracy-parameter ratio, that can be calculated with Eq.~\ref{eqn:parameter_ratio}. Taking into account this ratio, the best compromise between accuracy, loss and ratio is obtained with a \textbf{3x3 kernel}, and therefore we will use this one for the following tests. This is not really a surprise, as since VGG ~\cite{VGG} introduced the usage of this size of kernel, it has somehow become an standard. For example, two layers of a 3x3 kernel produce better results than one with a 5x5 kernel size.

\begin{equation}
    ratio=\frac{accuracy*10^{5}}{number\quad of \quad parameters}
    \label{eqn:parameter_ratio}
\end{equation}

\input{tables/baseline_limits/kernel_sizes}

\subsection{Number of filters}
Changing the kernel size we improve our ratio, but slightly worsen the accuracy. Hence more changes are needed. We tune the number of filters used on both convolutional layers, obtaining the results shown in Tab.~\ref{tab:filters_tuning}. Again the parameter that gives a better accuracy, 64 filters, reduces considerably the ratio.

Instead of using the same number of filters for both layers, we can combine two different number of filters. The best combination we found is using 64 filters for the first layer and 32 for the second, which performs with an accuracy of 0.78, 29480 parameters and a ratio of 2.64. This results improve the ratio from the baseline while maintaining the accuracy, thus we use this configuration from now on.

\input{tables/baseline_limits/num_filters}

\subsection{Activation functions}
The next parameter to be tuned is the activation function. In this case we can observe that the default activation function, \textbf{ReLU}, gives the best results as seen in Tab.\ref{tab:activation_tuning}. Therefore, changing the activation will not improve our system.
\input{tables/baseline_limits/activation_functions}

\subsection{Weight initialization}
\begin{figure}[t!]
\centering
\begin{tabular}{cc}
\includegraphics[height=2.3cm]{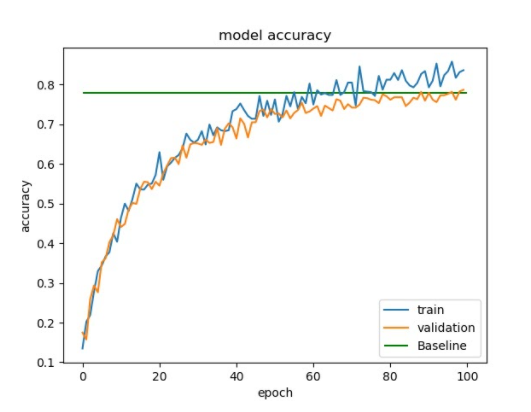} & 
\includegraphics[height=2.3cm]{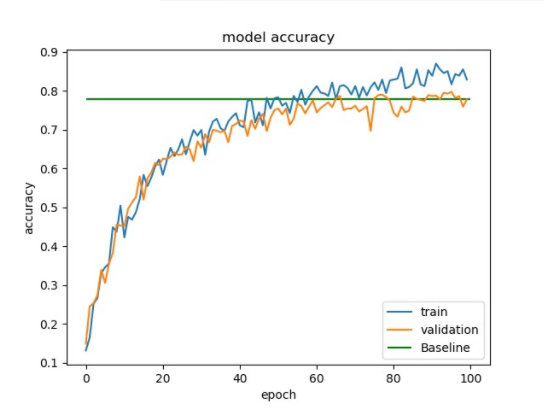} \\
(a) & (b) \\
\includegraphics[height=2.3cm]{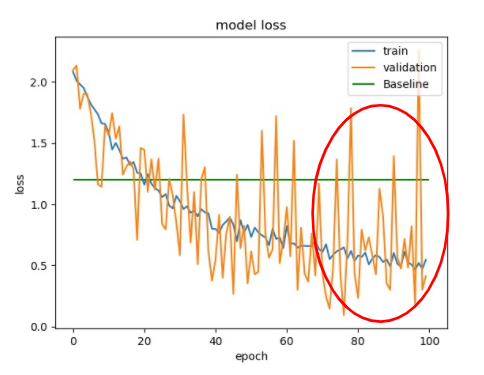} &
\includegraphics[height=2.3cm]{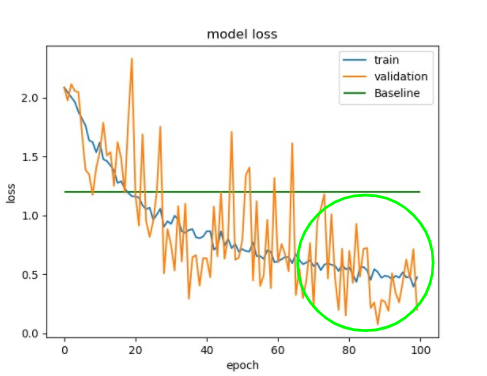} \\
(c) & (d)
\end{tabular}
\caption{Glorot (a) uniform and (b) normal accuracy curves; Glorot (c) uniform and (d) normal loss curves} \label{fig:initializations_tuning}
\end{figure}

Finally, we tune the weight initialization. Glorot, He and Random initialization are compared in Tab.~\ref{tab:initialization_tuning}. We add as well the pitfall: all zero initialization. We can empirically validate that what we should not initialize the weights to 0, as accuracy drops to 0.14. The reason is that by doing so, every neuron in the network computes the same output, the same gradients during backpropagation and undergo the exact same parameter updates. In other words, there is no source of asymmetry between neurons if their weights are initialized to be the same.
\input{tables/baseline_limits/baseline_weight}
Comparing the other initializations, numerically Glorot normal and uniform perform almost identically. To visualize the improvement in performance we need to check the accuracy and loss curves (Fig.~\ref{fig:initializations_tuning}). We can observe how with Glorot normal initialization, the loss curve (Fig.~\ref{fig:initializations_tuning}d) converges in a stable and smooth way. For this reason, \textbf{Glorot normal initialization} will be the one used.

Still our system did not improve its performance significantly, and we can state that continue tuning the baseline hyper-parameters is not the path to follow. 

\subsection{Adding depth}
Another option is to increase the depth of our architecture. The first approach is to keep adding blocks of 2D convolutional + 2d max pooling layers to our architecture. We use the hyperparameters obtained in the baseline tuning: 3x3 kernel size, Relu activation and Glorot normal initialization. However, in terms of number of filters we go back to using 32 of them for all the layers. In Tab.~\ref{tab:baseline_depth} we can observe how at first adding layers increases the accuracy up to 0.81, but when adding a fifth one the performance does not improve but even get worse. 

\begin{figure}[t!]
    \centering
    \includegraphics[height=0.7\textwidth]{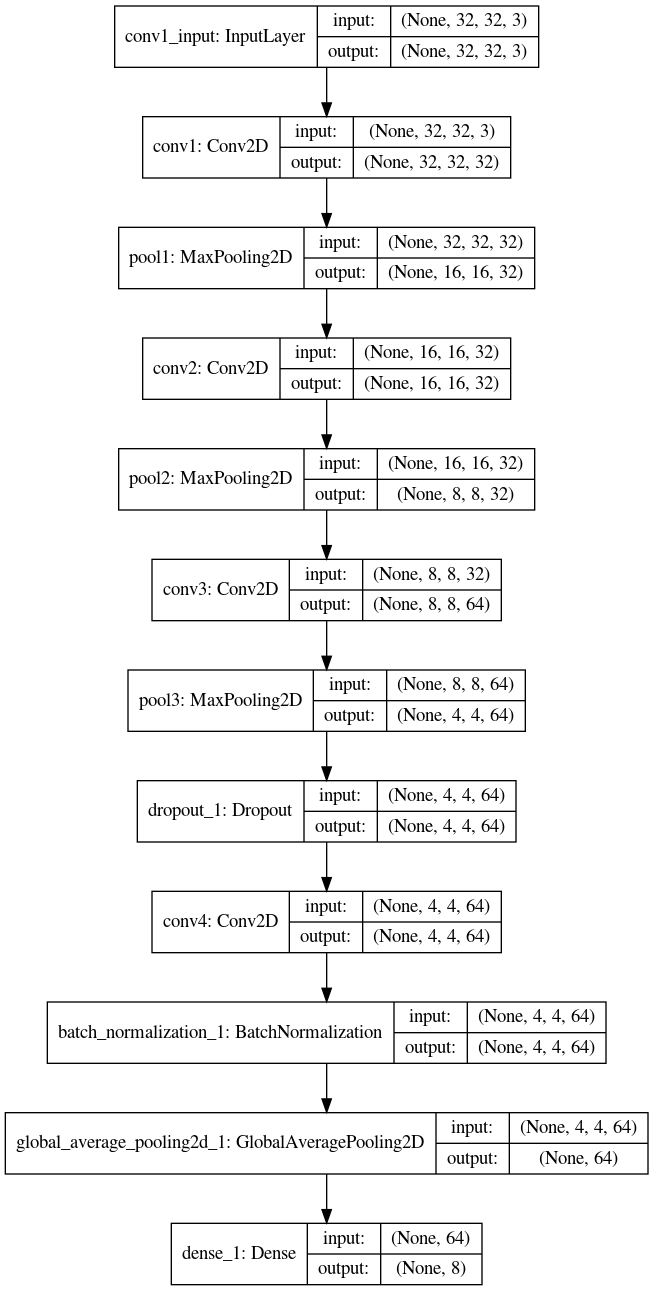} 
    \caption{Deep architecture} \label{fig:deep_model}
\end{figure}

Again adding depth randomly works but only up to a certain point. Additionally we only added one kind of layers, when there are a lot of possibilities to test. Using the fourth layer architecture, we add a dropout and a batch-norm layer, even though there is a consensus in not using them together. The final architecture is shown in Fig.~\ref{fig:deep_model}, and with it we get an accuracy of 0.82 with 66344 parameters and a ratio of 1.25.

\input{tables/adding_depth/adding_depth}

\subsection{Optimizer and Learning rate}
So far we ignored the hyper-parameters related to the optimizers but is another factor to take into account. Comparing the values obtained on Tab.~\ref{tab:baseline_optimizers} we can observe that Adam provides the best results.
\input{tables/adding_depth/optimizers.tex}

\begin{figure}[t!]
\centering
\begin{tabular}{cc}
\includegraphics[height=2.85cm]{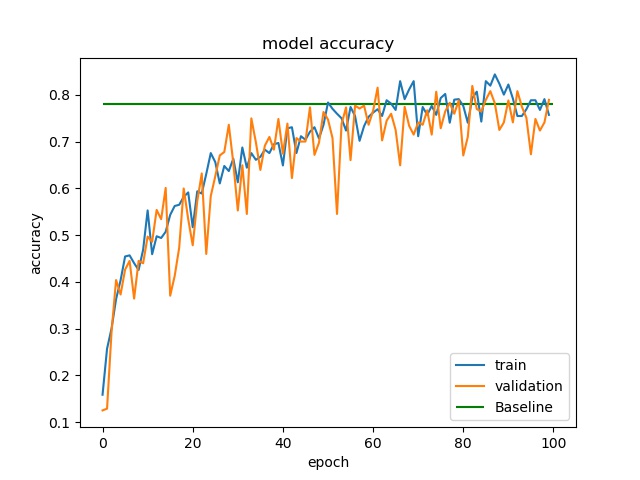} & 
\includegraphics[height=2.85cm]{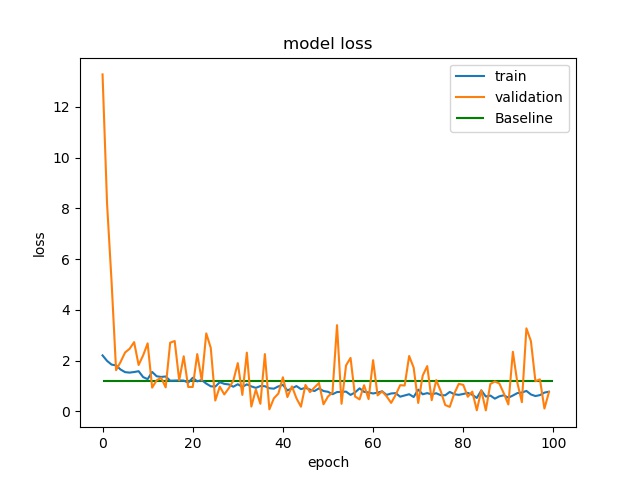} \\
(a) & (b) \\
\includegraphics[height=2.85cm]{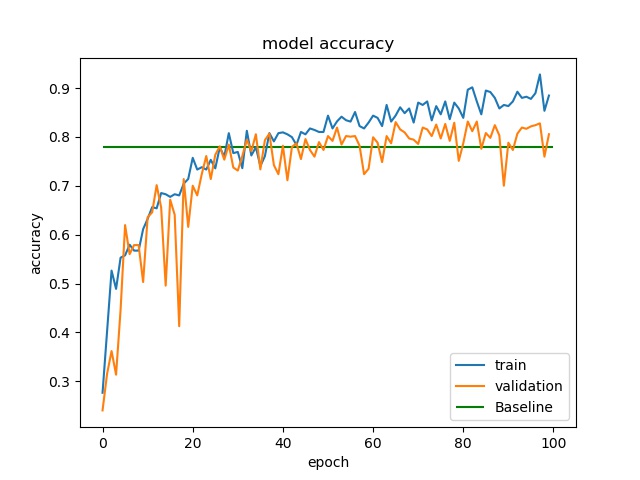} &
\includegraphics[height=2.85cm]{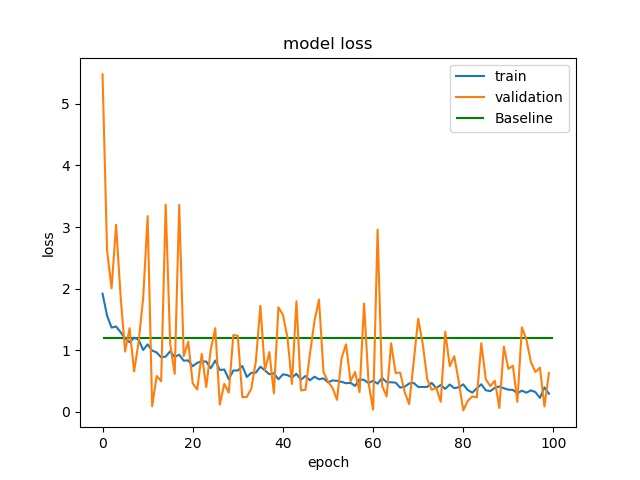} \\
(c) & (d)\\
\includegraphics[height=2.85cm]{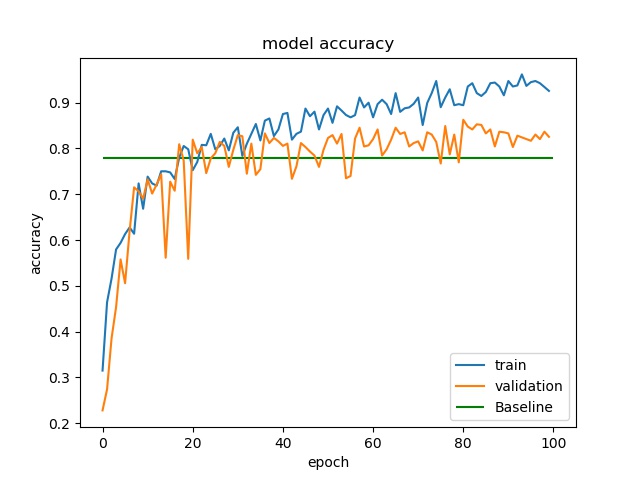} & 
\includegraphics[height=2.85cm]{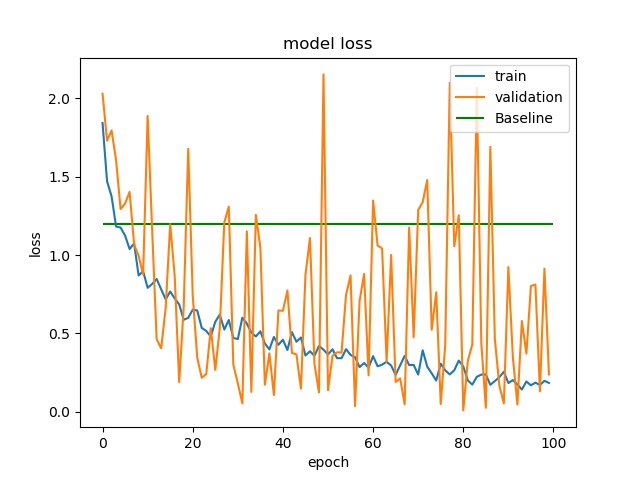} \\
(e) & (f) \\
\includegraphics[height=2.85cm]{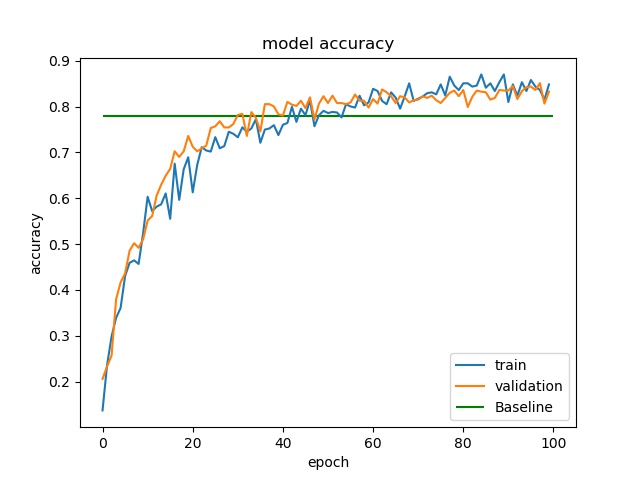} &
\includegraphics[height=2.85cm]{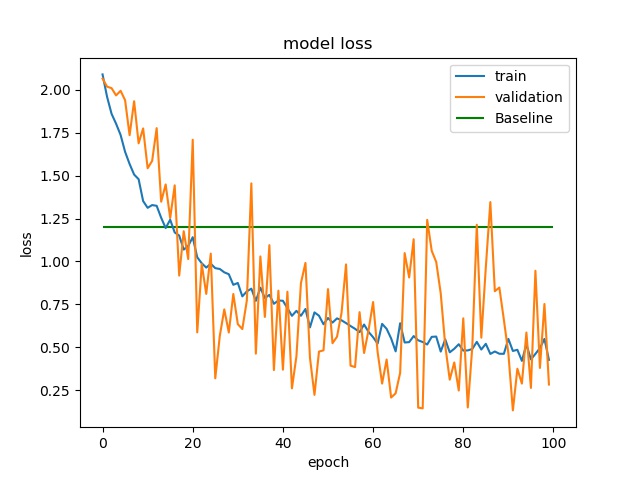} \\
(g) & (h)
\end{tabular}
\caption{Left column shows the accuracy curve and the right one the loss curve for learning rates of (a,b) 1e-1, (c,d) 1e-2, (e,f) 1e-3 and (g,h) 1e-4} \label{fig:lr_tuning}
\end{figure}

Strongly related to the optimizer, we have the learning rate, and it is usually hard to establish the best value. Numerically, we can observe in Tab.~\ref{tab:baseline_lr} that the smaller the learning rate the better results we obtain. But it is graphically that we can extract more information on how the system behaves. We can observe in Fig.~\ref{fig:lr_tuning} that the loss values get lower as the learning rate is decreased. In addition, for a learning rate of 1e-4, we observe that the accuracy curve (Fig.~\ref{fig:lr_tuning}g) has no overfitting.

\input{tables/adding_depth/learning_rates.tex}

\subsection{Input size}
It is not just the architecture of the system that affects the performance. How we preprocess the input data can play an important role too. In this case, we validate it by changing the size of the input images. We can observe that the performance changes significantly between the different sizes(Tab.~\ref{tab:baseline_input}), and it is with an input size of 64x64 that we obtain the best accuracy so far: 0.84. 
\input{tables/adding_depth/input_sizes.tex}

\subsection{Grad-CAM}
\begin{figure}[t!]
\centering
\begin{tabular}{cccc}
\includegraphics[height=3.85cm]{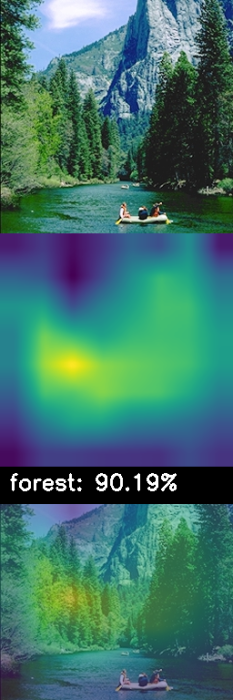} & 
\includegraphics[height=3.85cm]{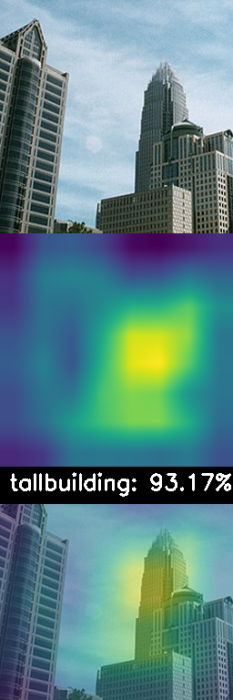} &
\includegraphics[height=3.85cm]{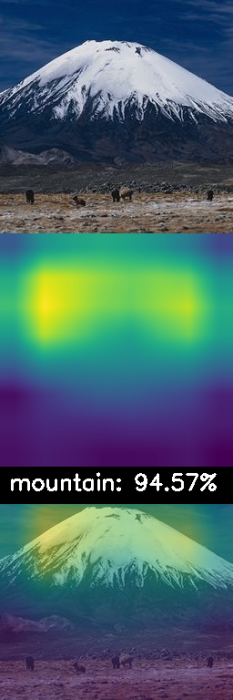} &
\includegraphics[height=3.85cm]{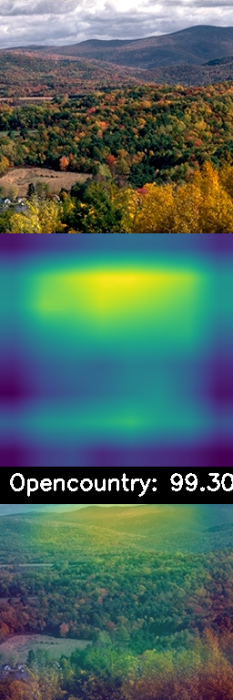} \\
(a) & (b) & (c) & (d)
\end{tabular}
\caption{Examples of activation maps for the (a) forest, (b) tall-building, (c) mountain and (d) open country classes} \label{fig:gradcam}
\end{figure}

It might feel that the deeper we go, the system becomes a darker box and it is harder to understand how every layer contributes our system. Fortunately, we can use techniques like Grad-CAM~\cite{GRADCAM} to visualize the regions of the input data that are more relevant for predicting an specific concept. In Fig.~\ref{fig:gradcam}a the activation for the forest class highlights the trees of the image. For the tall building class, in Fig.~\ref{fig:gradcam}b, it is clear that the highlighted object is the sky-scrapper. On the other hand, for the mountain example (Fig.~\ref{fig:gradcam}c) is the silhouette of the peak that allows the system to predict it correctly. Finally, we have an example of the open country class in Fig~\ref{fig:gradcam}d, which is, as we have seen before, the most difficult class to classify and there is not a distinctive activation to relate it to.

%% file: tables/baseline_limits/kernel_sizes.tex
\begin{table}[t!]
\centering
\caption{Tuning the kernel size}
\label{tab:kernel_tuning}
\begin{tabular}{{c|cccc}}
\toprule
Kernel Size & Accuracy & Loss & \# Parameters & Ratio \\ 
\midrule
\midrule
1x1 & 0.62 & 1.17 & 17576 & 3.55  \\
3x3 & 0.77 & 0.53 & 19368 & 3.97 \\
5x5 & 0.78 & 1.2 & 34472 & 2.2 \\
7x7 & 0.78 & 0.86 & 57256 & 1.37 \\
\bottomrule
\end{tabular}
\end{table}

%% file: tables/baseline_limits/num_filters.tex
\begin{table}[t!]
\centering
\caption{Tuning the number of filters}
\label{tab:filters_tuning}
\begin{tabular}{{c|cccc}}
\toprule
\# of Filters & Accuracy & Loss & \# Parameters & Ratio \\ 
\midrule
\midrule
8 & 0.71 & 0.63 & 3120 & 22.69  \\
16 & 0.76 & 0.58 & 7384 & 10.28 \\
32 & 0.77 & 0.53 & 19368 & 3.97 \\
64 & 0.79 & 0.35 & 57160 & 1.38 \\
128 & 0.76 & 0.51 & 188040 & 0.4 \\
\bottomrule
\end{tabular}
\end{table}

%% file: tables/baseline_limits/activation_functions.tex
\begin{table}[t!]
\centering
\caption{Tuning the activation functions}
\label{tab:activation_tuning}
\begin{tabular}{{c|cccc}}
\toprule
Activation & Accuracy & Loss & \# Parameters & Ratio \\ 
\midrule
\midrule
Relu & 0.78 & 0.56 & 29480 & 2.64  \\
Elu & 0.76 & 0.85 & 29480 & 2.58 \\
Tanh & 0.74 & 0.74 & 29480 & 2.5 \\
\bottomrule
\end{tabular}
\end{table}

%% file: tables/baseline_limits/baseline_weight.tex
\begin{table}[t!]
\centering
\caption{Tuning the weight initialization}
\label{tab:initialization_tuning}
\begin{tabular}{{c|cccc}}
\toprule
Initialization & Accuracy & Loss & \# Parameters & Ratio \\ 
\midrule
\midrule
Glorot Uniform & 0.78 & 0.56 & 29480 & 2.64  \\
Glorot Normal & 0.78 & 0.52 & 29480 & 2.64 \\
He Normal & 0.76 & 0.82 & 29480 & 2.57 \\
Random Normal & 0.77 & 0.8 & 29480 & 2.61 \\
Zeros & 0.14 & 2.1 & 29480 & 0.5 \\
\bottomrule
\end{tabular}
\end{table}

%% file: tables/adding_depth/adding_depth.tex
\begin{table}[t!]
\centering
\caption{Adding depth to the CNN}
\label{tab:baseline_depth}
\begin{tabular}{{c|cccc}}
\toprule
Architecture & Accuracy & Loss & \# Parameters & Ratio \\ 
\midrule
\midrule
Baseline & 0.77 & 0.53 & 19368 & 3.97 \\
Three layers & 0.8 & 0.75 & 20424 & 3.94 \\
Four layers & 0.81 & 0.8 & 29672 & 2.48 \\
Five layers & 0.77 & 0.67 & 38152 & 2.03 \\
\bottomrule
\end{tabular}
\end{table}

%% file: tables/adding_depth/optimizers.tex
\begin{table}[t!]
\centering
\caption{Tuning the optimizers}
\label{tab:baseline_optimizers}
\begin{tabular}{{c|cccc}}
\toprule
Optimizer & Accuracy & Loss & \# Parameters & Ratio \\ 
\midrule
\midrule
RMS prop & 0.74 & 0.52 & 66344 & 1.12\\
Adam & 0.82 & 0.68 & 66344 & 1.25 \\
SGD & 0.81 & 0.75 & 66344 & 1.22 \\
\bottomrule
\end{tabular}
\end{table}

%% file: tables/adding_depth/learning_rates.tex
\begin{table}[t!]
\centering
\caption{Tuning the learning rate}
\label{tab:baseline_lr}
\begin{tabular}{{c|cccc}}
\toprule
Learning Rate & Accuracy & Loss & \# Parameters & Ratio \\ 
\midrule
\midrule
1e-1 & 0.79 & 0.85 & 66344 & 1.19\\
1e-2 & 0.80 & 0.72 & 66344 & 1.21 \\
1e-3 & 0.82 & 0.64 & 66344 & 1.25 \\
1e-4 & 0.83 & 0.55 & 66344 & 1.25 \\
\bottomrule
\end{tabular}
\end{table}

%% file: tables/adding_depth/input_sizes.tex
\begin{table}[t!]
\centering
\caption{Tuning the input size}
\label{tab:baseline_input}
\begin{tabular}{{c|cccc}}
\toprule
Input Size & Accuracy & Loss & \# Parameters & Ratio \\ 
\midrule
\midrule
16 & 0.74 & 0.78 & 66344 & 1.12\\
32 & 0.83 & 0.55 & 66344 & 1.25 \\
64 & 0.84 & 0.56 & 66344 & 1.26 \\
128 & 0.79 & 0.63 & 66344 & 1.19 \\
\bottomrule
\end{tabular}
\end{table}

%% file: sections/03_method/aditya_part.tex
\section{Revisiting Weight Initialization}

\begin{figure}[t!]
\centering
\begin{tabular}{ccc}
\includegraphics[height=2.5cm]{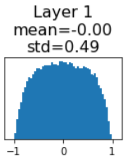} & 
\includegraphics[height=2.5cm]{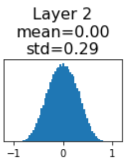} &
\includegraphics[height=2.5cm]{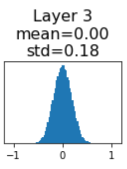} \\
(a) & (b) & (c) \\
\includegraphics[height=2.5cm]{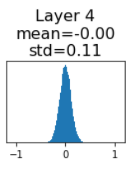} & 
\includegraphics[height=2.5cm]{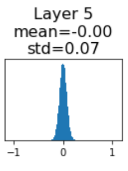} &
\includegraphics[height=2.5cm]{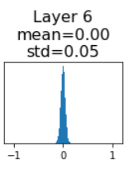} \\
(d) & (e) & (f) \\
\end{tabular}
\caption{Problems caused by poor initialization in neural networks. Network layers (a to f): 1 (shallow) to 6 (deeper). For deeper layers, activations start becoming zero and the gradients start collapsing to zero as well in the weights are not initialized properly} \label{fig:poor_init}
\end{figure}

\begin{figure}[t!]
    \centering
    \includegraphics[width=0.8\linewidth]{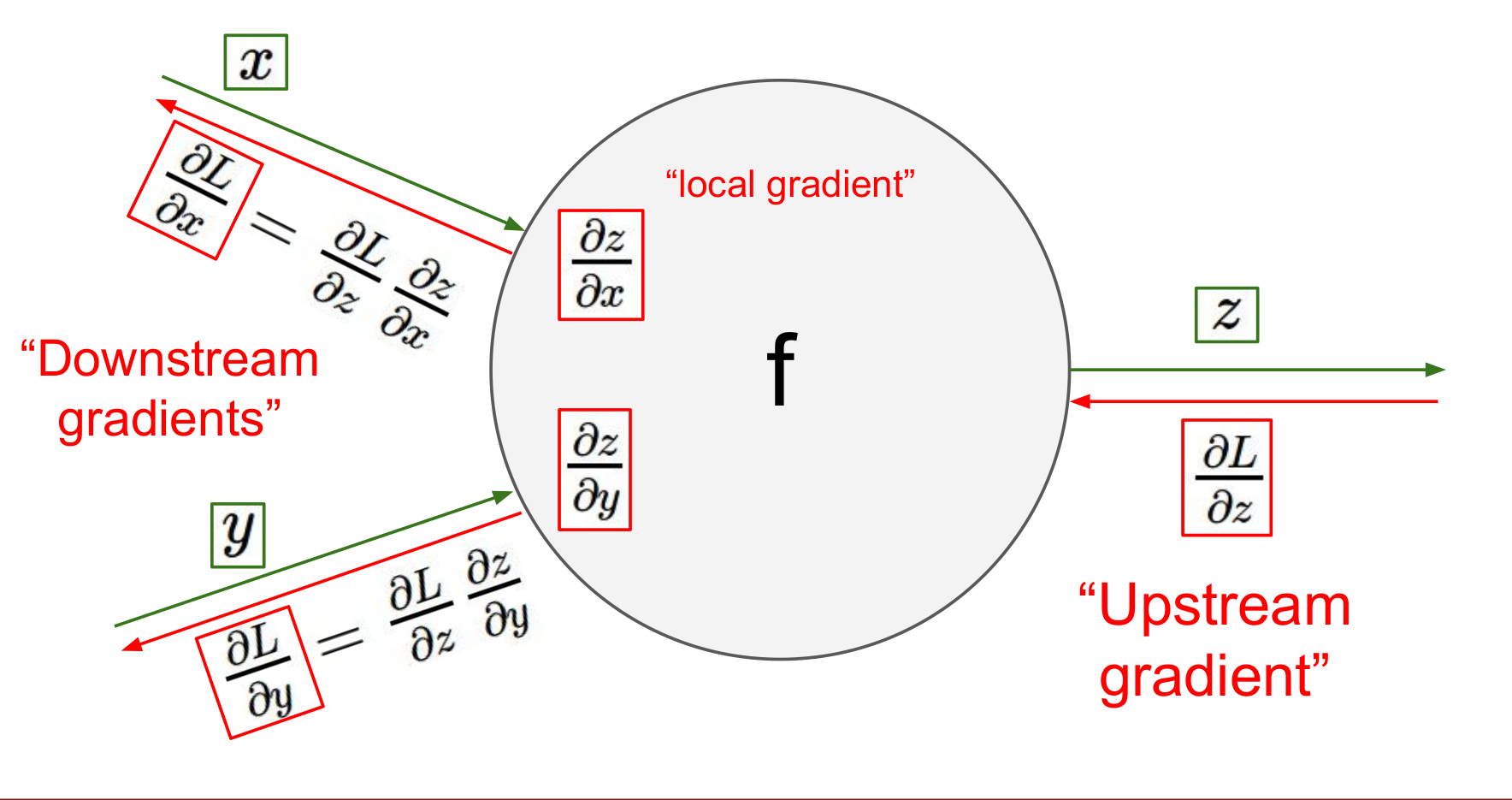} 
    \caption{Upstream gradients are multiplied by local gradients to get the downstream gradients during backprop}
    \label{fig:local_gradients_backprop}
\end{figure}

We want to answer the question of why a good initialization matters in neural networks? As Neural Networks involve a lot of matrix multiplications, the mean and variance of activations can quickly shoot off to very high values or drop down to zero as seen in Fig. \ref{fig:poor_init}. This will cause the local gradients of our layers to become NaN or zero and hence prevent our network from learning anything as the value of gradients depend on the forward activations as seen in Fig. \ref{fig:local_gradients_backprop}. A common strategy to avoid this is to initialize the weights of your network using the latest techniques. For example if you’re using ReLU activation after a layer, you must initialize your weights with Kaiming He initialization and set the biases to zero. This was introduced in the 2014 ImageNet winning paper \cite{resnet2014} from Microsoft. This ensures the mean and standard deviation of activations of all layers stay close to 0 and 1 respectively. 

\begin{figure}[t!]
    \centering
    \includegraphics[width=\linewidth]{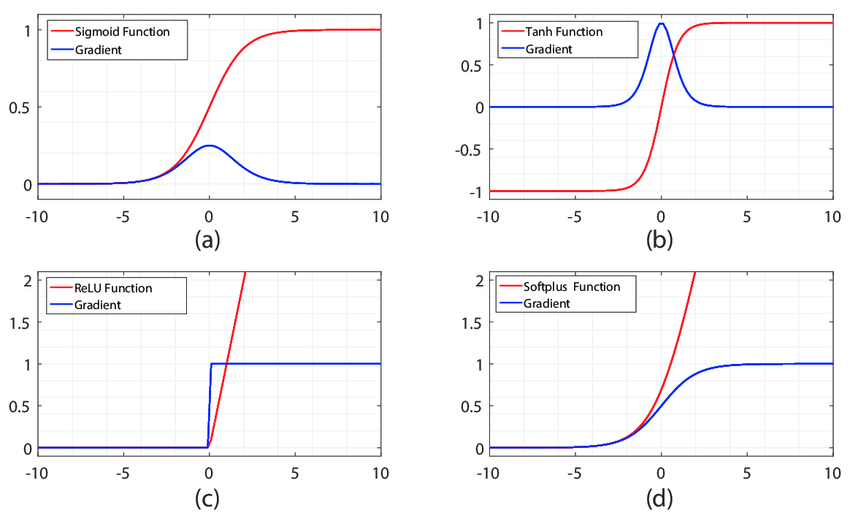} 
    \caption{Comparing the local gradient behavior of some common activation functions. (a)Sigmoid (b)Tanh (c)ReLU (d)Softplus} \label{fig:act_derivative}
\end{figure}

As you can see in Fig. \ref{fig:act_derivative},  the gradients in both sigmoid and tanh are non-zero only inside a certain range between [-5, 5]. Also notice that when using sigmoid, the local gradient achieves a maximum value of 0.25, thus every time gradient passes through a sigmoid layer, it gets diminished by at least 75 percent.

\section{Bringing Everything Together: TinyNet}
\subsection{Main Architecture}

\begin{figure}[t!]
    \centering
    \includegraphics[width=\linewidth]{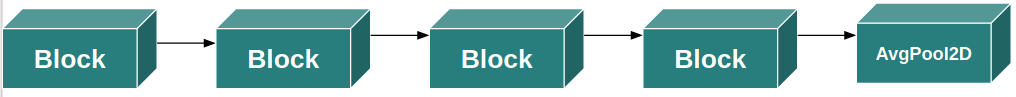} 
    \caption{Stacking blocks that make our TinyNet, followed by adaptive average pooling}
    \label{fig:block}
\end{figure}

We will be using the insight gained from all our previous experiments to train a 4-"block" CNN where each block may be composed of convolutions, activations, batchnorm, and residual connections. We will be using an input image size of 64x64, 3x3 kernels for each layer, stride 2 with padding 1 for downsampling (we will not be using max-pooling layers but only adaptive average pooling in the end layer). The outline for our TinyNet can be seen in Fig. \ref{fig:block}. The size of layers used are $[32, 64, 128,256].$

\begin{figure}[h]
    \centering
    \begin{tabular}{cc}
    \includegraphics[width=0.45\linewidth]{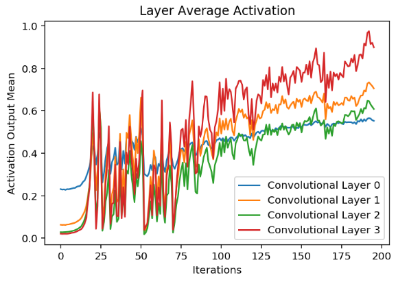} & 
    \includegraphics[width=0.45\linewidth]{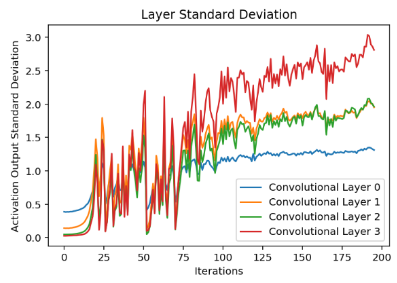} \\
    (a) & (b) \\
    \includegraphics[width=0.45\linewidth]{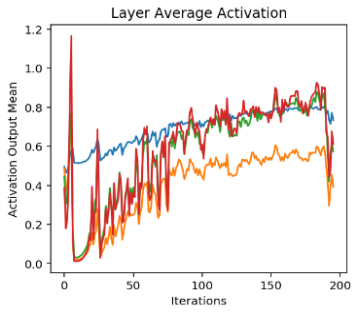} & 
    \includegraphics[width=0.45\linewidth]{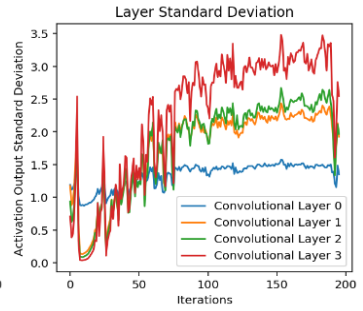} \\
    (c) & (d) \\
    \includegraphics[width=0.45\linewidth]{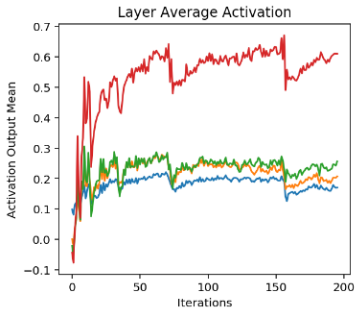} & 
    \includegraphics[width=0.45\linewidth]{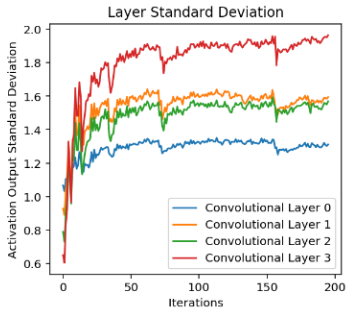} \\
    (e) & (f) \\
    \includegraphics[width=0.45\linewidth]{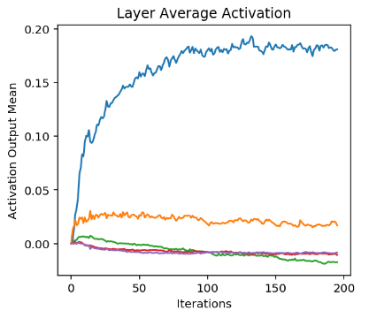} & 
    \includegraphics[width=0.45\linewidth]{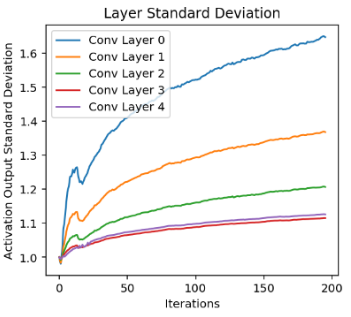} \\
    (g) & (h) \\
    \end{tabular}
    \caption{Layer activations: (left) mean and (right) standard deviation. Initializations: (a,b) Glorot uniform, (c,d) Kaiming He, (e,f) He + Leaky ReLU, and (g,h) He + LeakyReLU + BatchNorm}
    \label{fig:init_configurations}
\end{figure}

We track the performance of our network by keeping a track of the means and the variances of activations as our network trains. You can compare the network's performance for each of the block configuration we experiment with in Fig. \ref{fig:init_configurations} and compare their accuracies in Table \ref{tab:tinynet_main}. 

\begin{table}[]
    \centering
    \caption{Train and Test Accuracy for different "blocks" of our model}
    \label{tab:tinynet_main}
    \begin{tabular}{llcc}
    \toprule
    Model & Train & Test \\
    \midrule
        Keras Default (Glorot) & 0.842 & 0.813 \\
        Kaiming He init & 0.895 & 0.874 (+0.06)  \\
        He + Leaky ReLU & 0.887 & 0.883 (+0.07)\\
        He + Leaky ReLU + BatchNorm & 0.912 & 0.901 (+0.09)\\
    \bottomrule
    \end{tabular}
\end{table}

\begin{figure}[h]
    \centering
    \begin{tabular}{cc}
    \includegraphics[width=0.45\linewidth]{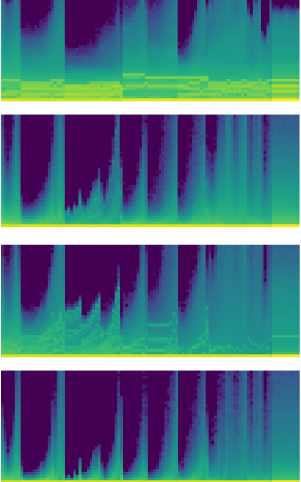} & 
    \includegraphics[width=0.45\linewidth]{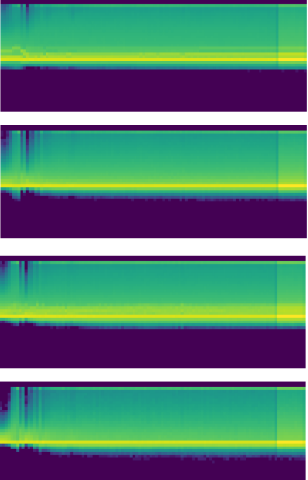} \\
    (a) & (b) \\
    \end{tabular}
    \caption{Histogram of Activations for each layer. From first layer(top) to last layer(bottom) (a) Keras Default (Glorot). (b) Kaiming He}
    \label{fig:hist}
\end{figure}

However these values are just aggregates of the layer parameters, so they don’t give us the full picture about how all the parameters are behaving. Rather than look at a single number we’d like to look at the distribution. To do that we can look at how the histogram of the parameters changes over time as shown in Fig. \ref{fig:hist}. The biggest concern is the amount of mass at the bottom of the histogram (at 0) in the original network. This is not good. In the last layer nearly 90 percent of the activations are actually 0. If you were training your model like this, it could appear like it was learning something, but you could be leaving a lot of performance on the table by wasting 90 percent of your activations. But, by using proper initiliazation and training techniques this can be fixed as we show. Notice in Fig.~\ref{fig:hist}b, it’s using the full richness of the possible activations and there’s not crashing of values.

Now that we know how to train our networks properly, let's see how it performs for different depth and width. We show the results in Table \ref{tab:tinynet_size}.

\begin{table}[]
    \centering
    \caption{Best configuration (Conv Block with Kaiming Initialization + Leaky ReLU + BatchNorm) used for these experiments.}
    \label{tab:tinynet_size}
    \begin{tabular}{clcccc}
    \toprule
    Layers & Filters & $\#$ Params & Train & Test & Ratio \\
    \midrule
        4 & $[32, 64, 128, 256]$ & 390,952& 0.912 & 0.901 & 0.23\\
        4 & $[16, 32, 64, 128]$ & 98,712 & 0.903 & 0.884 & 0.91\\
        3 & $[32, 64, 128]$     & 94,504 & 0.893 & 0.873 & 0.95\\
        3 & $[16, 32, 64]$      & 24,216 & 0.864 & 0.861 & 3.58\\
    \bottomrule
    \end{tabular}
\end{table}

\subsection{Adding Residual connections}
\begin{figure}[t!]
    \centering
    \includegraphics[width=0.6\linewidth]{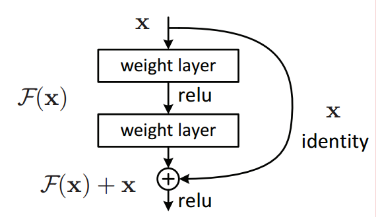} 
    \caption{Residual Connection}
    \label{fig:residual}
\end{figure}

\begin{figure}[t!]
    \centering
    \includegraphics[width=0.5\linewidth]{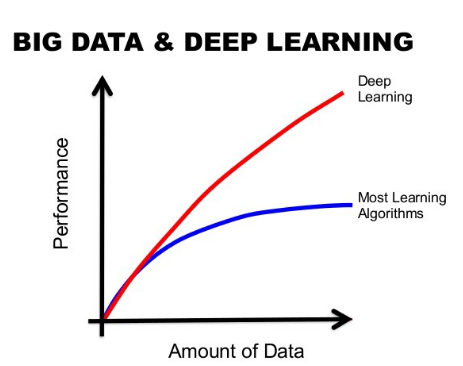} 
    \caption{Increased performance with increasing data for deep learning}
    \label{fig:bigdata}
\end{figure}

We added residual connections\cite{resnet2014} (Fig. \ref{fig:residual}) in our network but they didn't improve our accuracy at all, even though the number of parameters increased by 3 times. This was surprising for us, but the explanation is clear. As shown in Fig. \ref{fig:bigdata}, this time the bottleneck was not the model, but the amount of data that we had. With just 1800 training samples, it is difficult to train a good model from scratch. Even after using a lot of data augmentation, it wasn’t enough to train a deeper model.

Now that it was certain that making the model bigger won't be of much use, we shifted out attention to making the model smaller and more efficient. 

\subsection{Adding Depthwise Convolutions}

\begin{figure}[t!]
    \centering
    \begin{tabular}{cc}
    \includegraphics[width=0.45\linewidth]{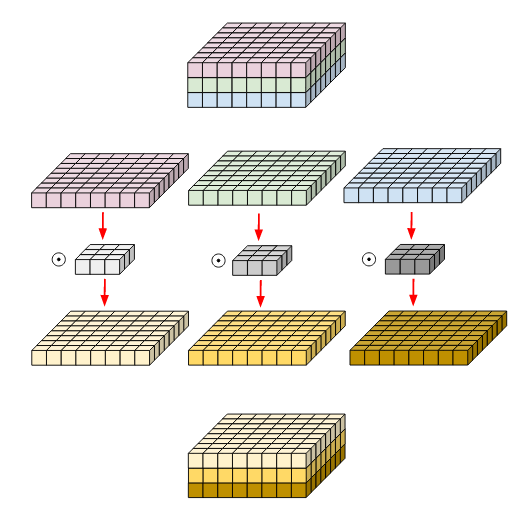} & 
    \includegraphics[width=0.45\linewidth]{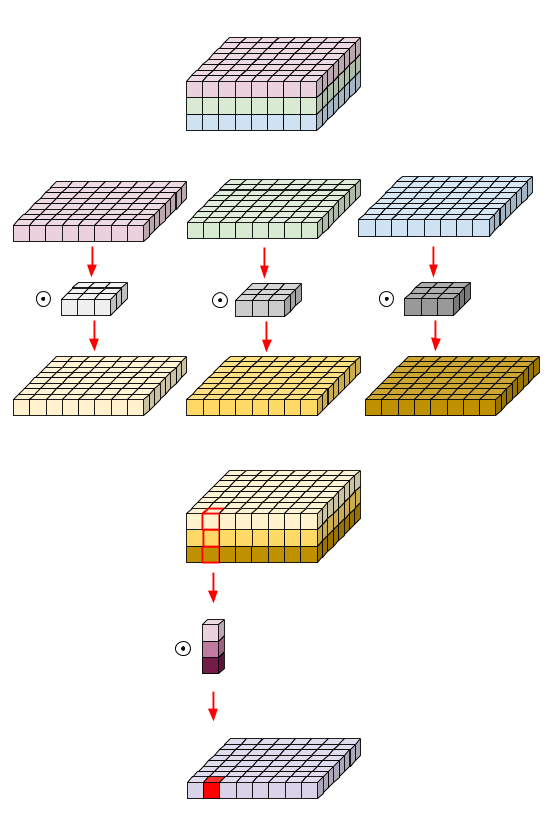} \\
    (a) & (b) \\
    \end{tabular}
    \caption{(a) Normal and (b) Depthwise convolution}
    \label{fig:normal_depth_conv}
\end{figure}

We replace normal convolutions in our model with depth-wise convolutions, adapting this idea from MobileNets \cite{mobilenet}. The MobileNet model is based on depthwise separable convolutions  which  is  a  form  of  factorized  convolutions which  factorize  a  standard  convolution  into  a  depthwise convolution and a 1×1 convolution called a pointwise convolution. This factorization has the effect of drastically reducing computation and model size. By expressing convolution as a two step process of filtering and combining, we get a reduction in computation per layer shown in Eq.~\ref{eq:computation_reduction}:

\begin{equation}
    reduction = \frac{1}{N} + \frac{1}{D_k^2}
    \label{eq:computation_reduction}
\end{equation}

where N is the number of kernels in the filter, and $D_k$ is the size of the kernel. For kernel size of 3x3, we can get approx. 9 times reduction in model size. A visual explanation of the difference is given in Fig. \ref{fig:normal_depth_conv}.

As we can see in the last row of the Table \ref{tab:tinynet_depthwise}, we are able to get an accuracy of 0.825 only with as low as 4,000 parameters.

\subsection{Optimizers}

We have used a few state-of-the-art tricks to automatically find the best learning rate and momentum for our Adam Optimizer. This includes the One Cycle Policy, \cite{onecyclepolicy} and Learning Rate Finder\cite{lrfinder} proposed by Leslie N. Smith that allowed us to train our networks at much higher learning rates, and thus they converge in lower number of epochs. Each epoch took only 3s to run and each of our models converged in less than 25 epochs, making the whole experiment around 1.5 mins. This was a big factor in our project as it allowed us to rapidly prototype and experiment different configurations and ideas.

\begin{table*}[]
    \centering
    \caption{Depthwise Convolutions replacing Normal Convolutions in the models presented in Table \ref{tab:tinynet_main}}
    \label{tab:tinynet_depthwise}
    \begin{tabular}{lclcccc}
    \toprule
    Type & Layers & Filters & $\#$ Params & Train & Test & Ratio \\
    \midrule
        Normal Conv & 4 & $[32, 64, 128, 256]$ & 390,952 & 0.912 & 0.901 & 0.23\\
        Depthwise Conv & 4 & $[32, 64, 128, 256]$ & 48,617 & 0.894 & 0.874 & 1.80\\
        
        Normal Conv & 4 & $[16, 32, 64, 128]$ & 98,712 & 0.903 & 0.884 & 0.91\\
        Depthwise Conv & 4 & $[16, 32, 64, 128]$ & 13,577 & 0.873 & 0.863 & 6.4\\
        
        Normal Conv & 3 & $[16, 32, 64]$      & 24,216 & 0.864 & 0.861 & 3.58\\
        Depthwise Conv & 3 & $[16, 32, 64]$      & 3,913 &  0.836 & 0.825 & 21.08\\
    \bottomrule
    \end{tabular}
\end{table*}

%% file: sections/conclusions.tex
\section{CONCLUSIONS}
\label{sec:conclusions}
In this paper, we have explored some traditional and Deep Learning techniques that can be used in a classification system, and we have seen the advantages and disadvantages of using some models compared to others. From the results obtained, we can draw the following conclusions. 

First of all, classic approaches (e.g. BoVW) can provide good results, but they are limited and would not make a reliable nor robust system for a real world application. However, it still performs better than a simple Deep Learning technique like the Multi-Layer Perceptron. 
MLPs are too simple for our image classification problem, and even when optimizing their hyper-parameters the performance is poor. Using the MLP system to extract deep features to use afterwards on SVM, or as descriptors for the BoVW system does not provide us with better results neither. 

Fine-tuning on a pre-trained network (e.g.  InceptionV3) gives the best results in terms of accuracy when unfreezing and retraining the weights, 96 percent. But if we take into account the number of parameters, it is not an efficient system. Indeed, we can remove layers and reduce by five times the amount of parameters without losing accuracy, and still be an overkill for our specific dataset. 

When training a model from scratch, a good weight initialization is important for making sure our models train properly. We need a good amount of data for training deep neural networks, and with the dataset that we had (1,800 samples) we could not improve our accuracy above a threshold (90 percent), because the data wasn't enough to learn features that were representative enough of the dataset. Still our model is better fitted to the specific problem and dataset: while the pruned InceptionV3 had 5M parameters, in our model we have around 4K parameters.